\documentclass[lettersize,journal]{IEEEtran}
\usepackage{amsmath,amsfonts,amssymb}
\usepackage{algorithmic}
\usepackage{algorithm}
\usepackage{array}
\usepackage[caption=false,font=normalsize,labelfont=sf,textfont=sf]{subfig}
\usepackage{textcomp}
\usepackage{stfloats}
\usepackage{url}
\usepackage{verbatim}
\usepackage{graphicx}
\usepackage{cite}
\usepackage{booktabs}
\usepackage{multirow}
\usepackage{threeparttable}
\usepackage{orcidlink}
\usepackage{tikz}
\usetikzlibrary{shapes.geometric,arrows.meta,positioning,backgrounds,calc}
\usepackage{pgfplots}
\pgfplotsset{compat=1.18}
\begin{document}

\title{SkillGraph: Graph Foundation Priors for LLM Agent Tool Sequence Recommendation}

\author{Hao Liu$^{\orcidlink{0009-0004-7614-9597}}$
and Dongyu Li$^{\orcidlink{0000-0001-8338-0536}}$%
\thanks{Hao Liu is with the Hangzhou International Innovation Institute, Beihang University, Hangzhou 311115, China (e-mail: lh@computer.org).}%
\thanks{Dongyu Li is with the School of Cyber Science and Technology, Beihang University, Beijing 100191, China (e-mail: dongyuli@buaa.edu.cn).}
}

\maketitle

% ============================================================
\begin{abstract}
% ============================================================
LLM agents must select tools from large API libraries and order them correctly. Existing methods use semantic similarity for both retrieval and ordering, but ordering depends on inter-tool data dependencies that are absent from tool descriptions. As a result, semantic-only methods can produce \emph{negative} Kendall-$\tau$ in structured workflow domains. We introduce \textbf{SkillGraph}, a directed weighted execution-transition graph mined from 49,831 successful LLM agent trajectories, which encodes workflow-precedence regularities as a reusable graph foundation prior.
Building on this graph foundation prior, we propose a \textbf{two-stage decoupled framework}: GS-Hybrid retrieval for candidate selection and a learned pairwise reranker for ordering. On ToolBench (9,965 test instances; $\sim$16,000 tools), the method reaches Set-F1\,=\,0.271 and Kendall-$\tau$\,=\,0.096; on API-Bank, Kendall-$\tau$ improves from $-0.433$ to $+0.613$.
Under identical Stage-1 inputs, the learned reranker also outperforms LLaMA-3.1-8B Stage-2 rerankers.
\end{abstract}


% ============================================================
\section{Introduction}
\label{sec:intro}
% ============================================================

\IEEEPARstart{T}{ool}-augmented large language model (LLM) agents now face a straightforward-sounding yet increasingly thorny planning problem: when a user raises a query and the agent can reach thousands of APIs, which tools should it call and in what order?
This recommendation step largely determines agent performance in domains that stretch from workflow automation to scientific data analysis~\cite{react,toolllm,toolformer}, yet human-crafted plans or prompt engineering alone collapse once tool libraries swell to the tens of thousands~\cite{toolllm}.
At that scale, we need automated recommenders that can quickly assemble the right tool chains even for queries the agent has never encountered.

Most existing approaches treat tool recommendation as a single retrieval problem: embed the query, rank tools by description similarity, and read execution order from those scores or from co-occurrence graphs~\cite{toolllm}.
This collapses two structurally different problems into one.
(i)~\emph{Tool set selection} is a relevance problem---query and tool descriptions share vocabulary, so embedding similarity works.
(ii)~\emph{Tool sequence ordering} is a dependency problem---the correct execution order depends on which tool's outputs feed subsequent steps, and that information is absent from tool descriptions.

\begin{figure}[!t]
\centering
\includegraphics[
  width=\columnwidth,
  trim=117.98bp 1061.36bp 648.92bp 98.99bp,
  clip
]{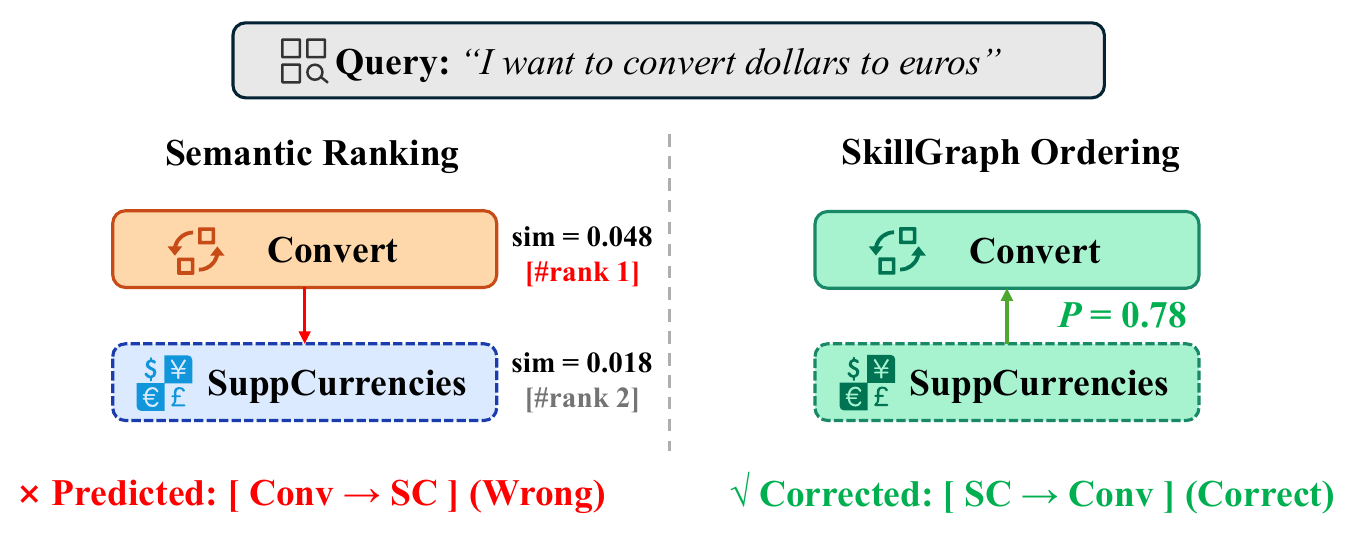}
\caption{The \emph{selection--ordering gap}. For query ``I want to convert dollars
to euros'', semantic similarity ranks \texttt{Convert} first
(sim\,=\,0.048\,$>$\,0.018), inverting the required execution order:
one must first call \texttt{SuppCurrencies} (short for
\texttt{supported\_currencies\_for\_currency\_converter\_v2}) to obtain
the valid currency list before invoking \texttt{Convert}.
SkillGraph, mined from LLM agent trajectories, encodes the dependency
\texttt{SC}$\!\to\!$\texttt{Conv} ($P=0.78$) and restores the correct sequence.}
\label{fig:motivation}
\end{figure}

We call this problem the \emph{selection-ordering signal gap} and characterize it empirically.
On API-Bank~\cite{apibank}, a structured multi-step workflow benchmark, \emph{all} semantic-based methods produce negative Kendall-$\tau$ correlation with ground-truth tool sequences (as low as $-0.433$): sorting tools by semantic relevance actively inverts the correct execution order.
Tool execution dependencies are not encoded in tool names or descriptions, and cannot be inferred from descriptions alone.
Consider the query \emph{``I want to convert dollars to euros.''} Semantic similarity ranks \texttt{Convert} first (sim\,=\,0.048) because it directly matches the query intent; yet \texttt{SuppCurrencies} \emph{must} run first to retrieve the valid currency list that \texttt{Convert} requires (see Fig.~\ref{fig:motivation}).
This dependency is invisible in descriptions alone---it only surfaces in trajectory records where successful agents consistently invoke \texttt{SuppCurrencies} before \texttt{Convert}.

We mine these dependency patterns at scale into a graph structure, \textbf{SkillGraph}, a graph foundation prior~\cite{gfm_survey} encoding workflow-precedence cues extracted from large-scale agent experience.
SkillGraph is built once from 49,831 successful ToolBench trajectories and reused across queries, analogously to how language foundation models are pretrained once and applied broadly.
The graph exhibits strong community structure (modularity~$= 0.892$, mean community purity~$= 0.779$, NMI~$= 0.695$), confirming that the mined graph captures semantically coherent tool workflow clusters rather than spurious co-occurrence noise.

Building on SkillGraph, we propose a two-stage decoupled framework: Stage~1 uses GS-Hybrid retrieval to build a candidate tool set, and Stage~2 applies a learned pairwise reranker to order that fixed set.
Decoupling prevents the ordering objective from degrading selection quality.
On ToolBench, LR is the sole Pareto-optimal method; on API-Bank, it raises Kendall-$\tau$ from $-0.433$ to $+0.613$.
Under the same Stage-1 input, LR also outperforms LLaMA-3.1-8B as a Stage-2 reranker.

We make four contributions:
\begin{enumerate}

\item \textbf{Selection-Ordering Signal Gap.}
We identify and quantify an asymmetry between tool selection and ordering: semantic methods achieve adequate selection quality but produce negative ordering correlation in structured workflow domains---a failure mode not previously characterized in the tool recommendation literature.

\item \textbf{SkillGraph.}
We construct a directed weighted execution-transition graph from large-scale LLM agent trajectories (ToolBench) and show that SkillGraph encodes semantically coherent workflow-precedence cues that complement semantic embeddings.

\item \textbf{Two-Stage Decoupled Framework.}
We propose a decoupled framework that assigns each sub-problem its appropriate signal: hybrid graph-semantic retrieval for selection and a learned pairwise reranker with SkillGraph features for ordering. The framework achieves Pareto-optimal performance across both sub-tasks.

\item \textbf{Empirical Validation.}
We evaluate on two benchmarks against semantic, graph-based, and LLM baselines, using bootstrapped significance tests and a low-resource second-benchmark validation. LR transfers to API-Bank without retraining, confirming that the learned dependency prior generalizes beyond ToolBench.

\end{enumerate}

% ============================================================
\section{Related Work}
\label{sec:related}
% ============================================================

\subsection{Tool Learning for LLM Agents}

Recent work equips LLMs with external tool use~\cite{toolsurvey,toolsurvey2025}.
Toolformer~\cite{toolformer} fine-tunes models to self-annotate API calls in text for single-step tool invocation.
ToolLLM~\cite{toolllm} scales this to 16,000+ real-world APIs via depth-first search tree (DFSDT) planning and contributes the ToolBench dataset used in this work.
API-Bank~\cite{apibank} evaluates tool-augmented LLM capabilities across multiple difficulty levels.
ReAct~\cite{react} interleaves chain-of-thought reasoning with tool calls, improving sequential decision-making in agentic settings.
These works train LLMs to \emph{invoke} tools correctly; none addresses \emph{recommending} the right tool sequence from a large library for a new query.
Our work treats tool sequence prediction as a recommendation problem---separating retrieval from ordering---rather than a generation or fine-tuning problem.

\subsection{Retrieval-Augmented Planning and API Retrieval}

Retrieval-augmented generation (RAG) has been adapted to agent planning by retrieving relevant tools, documents, or demonstrations at query time~\cite{rag}.
Semantic embedding retrieval is the dominant paradigm~\cite{sentencetransformers,dpr,huggingface}: tools are indexed by their description embeddings, and the top-$K$ matches are returned for a query.
Some systems augment retrieval with co-occurrence signals or knowledge graphs to capture tool relationships~\cite{toolllm}.
All these approaches use retrieval scores to simultaneously determine tool identity and order, conflating selection and ordering in a way that hurts ordering quality; decoupling the two stages is the direct fix.
Outside LLM agents, software engineering has studied API usage mining and invocation-order regularities from code repositories, including MAPO~\cite{mapo} and subsequent pattern-mining systems for Android APIs~\cite{androidpatterns}.
These works mine reusable call patterns for developer assistance but are not query-conditioned recommenders and do not address tool-set selection or sequence ordering for open-ended LLM agent planning.

\subsection{Sequential Recommendation}

Sequential recommendation~\cite{sasrec} models user interaction sequences to predict the next item.
SASRec~\cite{sasrec} applies self-attention over interaction histories to capture temporal preferences.
Recent studies have further enriched sequential recommendation with dynamic graph modeling, graph-based embedding smoothing, and counterfactual data augmentation~\cite{dgnn_seqrec_tkde,gbes_seqrec_tkde,cf_seqrec_tkde}.
Tool sequences differ from user-item sequences: rather than reflecting evolving user \emph{preferences}, they are governed by \emph{functional dependencies}---structural constraints on how data flows between tools.
These constraints do not shift over time and cannot be inferred from user history; they are fixed properties of tool interfaces, visible only in successful execution trajectories.
SkillGraph encodes these functional dependency constraints, which separates it from preference-based sequential recommendation.

\subsection{Learning to Rank}

Pairwise learning-to-rank methods~\cite{ranknet} train a model to predict the relative order of item pairs, then aggregate pairwise preferences into a total ranking.
RankNet~\cite{ranknet} introduced the pairwise cross-entropy loss used in our Learned Reranker.
Listwise methods later optimized ranking quality at the whole-list level rather than through pairwise preferences alone~\cite{listwise}.
We do not propose a new ranking architecture; we identify SkillGraph-derived transition features as the key input signal for tool ordering.
Prior work applying pairwise ranking to tool or API recommendation is sparse, and standard ranking features (relevance scores, position statistics) alone are insufficient for ordering---as the LR feature ablation shows.

\subsection{Graph Foundation Models}

Graph foundation models (GFMs)~\cite{gfm_survey} pre-train general-purpose graph representations that transfer across diverse graph domains, analogously to language foundation models.
LLMs have been applied to graph-structured data~\cite{gfm_survey}, and GNNs pre-trained on large corpora transfer across graph domains.
SkillGraph is a \emph{domain-specific graph foundation prior}: built once from large-scale trajectory data, it applies to arbitrary queries without retraining.
Unlike GNN-based GFMs, which produce node embeddings that require a separate inference model, SkillGraph encodes its prior directly as edge transition weights---interpretable, lightweight, and immediately usable in both retrieval (Stage~1) and ranking (Stage~2) with no additional graph inference at test time.
This design sidesteps the scalability costs of running GNN inference over a 4,988-node, 39,034-edge graph for every query.

% ============================================================
\section{Problem Formulation}
\label{sec:problem}
% ============================================================

\subsection{Notation and Task Definition}

We denote the tool library as $\mathcal{T} = \{t_1, t_2, \ldots, t_N\}$, with $N \approx 16{,}000$ tools in ToolBench.
Each tool $t_i$ carries a natural-language description $d_i$ covering its name, function signature, and intended use.
A query $q$ specifies a task requiring several coordinated tool calls.
We collect training signal from agent execution logs: the trajectory dataset $\mathcal{D} = \{(q^{(i)}, S^{*(i)})\}_{i=1}^{M}$ pairs each query with the ordered tool sequence $S^{*(i)} = [t_{\pi(1)}^*, t_{\pi(2)}^*, \ldots, t_{\pi(L^{*(i)})}^*]$ that a successful agent actually invoked, where $L^{*(i)} = |S^{*(i)}|$ is the ground-truth sequence length.

\textbf{Task.} Given $q$ and $\mathcal{T}$, produce an ordered sequence
\begin{equation}
\hat{S} = [\hat{t}_1, \hat{t}_2, \ldots, \hat{t}_{K_{\mathrm{eval}}}]
\label{eq:task}
\end{equation}
that matches $S^*$ in both which tools appear and in what order they execute.

\subsection{Decomposition into Two Sub-Problems}

Tool sequence recommendation resists a single unified objective: the two sub-problems draw on different signals, and optimizing one does not improve the other.

\textbf{Sub-problem 1: Tool Set Selection.}
Find the $K_{\mathrm{eval}}$-element subset $\hat{\mathcal{S}} \subseteq \mathcal{T}$ that maximally overlaps with $\mathcal{S}^* = \{t_1^*, \ldots, t_{L^*}^*\}$ (the unordered ground-truth tool set; cf.\ $S^*$ for the ordered sequence).
This reduces to relevance matching---does the tool description fit the query intent?

\textbf{Sub-problem 2: Sequence Ordering.}
Given $\hat{\mathcal{S}}$, find the permutation $\sigma: \{1,\ldots,K_{\mathrm{eval}}\} \to \{1,\ldots,K_{\mathrm{eval}}\}$ whose output $[\hat{t}_{\sigma(1)}, \ldots, \hat{t}_{\sigma(K_{\mathrm{eval}})}]$ best recovers $S^*$.
This requires modeling execution dependencies---which tool's output feeds the next step.
Such dependencies appear nowhere in tool descriptions; they are recoverable only from execution trajectories.

The gap between these signals motivates our two-stage design (Section~\ref{sec:framework}): Stage~1 retrieves candidates by relevance; Stage~2 reorders them by learned dependency patterns.

\subsection{Evaluation Metrics}

Selection and ordering require separate metrics because a method that retrieves the right tools may still sequence them incorrectly, and vice versa.

\textbf{Selection metrics.}
Let $\hat{\mathcal{S}}$ and $\mathcal{S}^*$ denote the predicted and ground-truth tool sets:
\begin{align}
\text{Set-Prec} &= \frac{|\hat{\mathcal{S}} \cap \mathcal{S}^*|}{|\hat{\mathcal{S}}|}, \quad
\text{Set-Recall} = \frac{|\hat{\mathcal{S}} \cap \mathcal{S}^*|}{|\mathcal{S}^*|}, \label{eq:sel_pr} \\
\text{Set-F1} &= \frac{2 \cdot \text{Set-Prec} \cdot \text{Set-Recall}}{\text{Set-Prec} + \text{Set-Recall}}. \label{eq:f1}
\end{align}

\textbf{Ordering metrics.}
Let $\hat{S} = [\hat{t}_1, \ldots, \hat{t}_{K_{\mathrm{eval}}}]$ and $S^* = [t_1^*, \ldots, t_{L^*}^*]$, where $L^* = |S^*|$.
We use four measures that capture different facets of ordering quality.

\emph{Ordered Precision} (Ord.Prec) measures pairwise order accuracy on the common tool subset $\mathcal{C} = \hat{\mathcal{S}} \cap \mathcal{S}^*$.
It asks: among all unordered pairs of correctly retrieved tools, what fraction appear in the same relative order as in the ground truth?
\begin{equation}
\text{Ord.Prec}
= \frac{|\text{concordant pairs}|}{\binom{|\mathcal{C}|}{2}}.
\label{eq:ord_prec}
\end{equation}
When $|\mathcal{C}| < 2$, we define $\text{Ord.Prec} = 0$.

\emph{Kendall-$\tau$} measures rank correlation on the common tool subset $\mathcal{C} = \hat{\mathcal{S}} \cap \mathcal{S}^*$, and is invariant to set errors.
Let $r_{\hat{S}}(t)$ and $r_{S^*}(t)$ denote the position of tool $t$ in $\hat{S}$ and $S^*$, respectively.
A pair $(t_a, t_b) \in \mathcal{C}^2$ is \emph{concordant} if $r_{\hat{S}}(t_a) < r_{\hat{S}}(t_b)$ iff $r_{S^*}(t_a) < r_{S^*}(t_b)$, and \emph{discordant} otherwise:
\begin{equation}
\tau = \frac{|\text{concordant pairs}| - |\text{discordant pairs}|}{\binom{|\mathcal{C}|}{2}}.
\label{eq:kendall}
\end{equation}
$\tau \in [-1, 1]$, where $\tau = 1$ is perfect order agreement and $\tau < 0$ indicates systematic order reversal; when $|\mathcal{C}| < 2$, we define $\tau = 0$.

\emph{Transition Accuracy} (Trans.Acc) measures the fraction of ground-truth consecutive pairs $(t_i^*, t_{i+1}^*)$ for which the successor appears shortly after its predecessor in the prediction, reflecting whether the model preserves local execution steps:
\begin{equation}
\text{Trans.Acc} =
\frac{|\{(t_i^*, t_{i+1}^*) : 0 < r_{\hat{S}}(t_{i+1}^*) - r_{\hat{S}}(t_i^*) \le 2\}|}{L^* - 1}.
\label{eq:transacc}
\end{equation}
When $L^* < 2$, we define $\text{Trans.Acc} = 0$.

\emph{First-Tool Accuracy} (1st.Acc) $= \mathbf{1}[\hat{t}_1 = t_1^*]$ indicates whether the entry-point tool is correct.
This matters because the first tool determines the context all later tools receive.

No single metric is sufficient: high Kendall-$\tau$ with low 1st.Acc captures global order well while misidentifying the entry point; high 1st.Acc with low Trans.Acc gets the start right but loses track of dependencies mid-sequence.

\subsection{Oracle-$K$ Protocol}

All experiments reveal the ground-truth sequence length $L^* = |S^*|$ at evaluation time and set the prediction budget to
\begin{equation}
K_{\mathrm{eval}} = \max(L^*, 3).
\end{equation}
Methods therefore return up to $K_{\mathrm{eval}}$ tools, while the gold sequence remains length $L^*$.
When $L^* < 3$, extra predicted tools are treated as over-predictions in the set and ordering metrics.
This isolates selection and ordering quality from sequence-length estimation errors, enabling clean comparisons across methods.
The trade-off is that real deployment requires inferring $K_{\mathrm{eval}}$ from the query itself---a capability none of our methods possess.
We treat adaptive length prediction as orthogonal and exclude it from this work.

% ============================================================
\section{SkillGraph Construction}
\label{sec:skillgraph}
% ============================================================

SkillGraph is a directed weighted graph $\mathcal{G} = (\mathcal{V}, \mathcal{E}, \mathbf{W})$
mined from successful LLM agent trajectories.
Nodes $\mathcal{V}$ represent tools; directed edges $\mathcal{E}$ represent observed execution transitions; edge weights $\mathbf{W}$ encode conditional transition probabilities.
Built once from training data, SkillGraph captures workflow-precedence cues and is reused across all future queries.

\subsection{Trajectory Data Processing}

We build SkillGraph from the ToolBench training corpus~\cite{toolllm}, which contains LLM agent trajectories collected via depth-first search tree (DFSDT) planning.
We retain only \emph{successful} trajectories---those where the agent completed the task and received a positive reward---yielding $M = 49{,}831$ trajectory-query pairs.

Each trajectory is parsed into an ordered tool invocation sequence:
\begin{equation}
    \tau = [t_{\pi(1)},\, t_{\pi(2)},\, \ldots,\, t_{\pi(L)}],
    \label{eq:traj}
\end{equation}
where each tool $t_{\pi(i)} \in \mathcal{T}$ is identified by its canonical identifier (API name + tool name) to ensure uniqueness across the $\sim$16,000-tool library.
Tool calls within a trajectory are deduplicated while preserving order, as repeated invocations of the same tool carry no additional dependency signal.
The resulting dataset $\{(q^{(i)}, \tau^{(i)})\}_{i=1}^{M}$ is used exclusively for graph construction and reranker training; test queries are held out.

\subsection{Graph Construction}

\textbf{Nodes.}
Each unique tool that appears at least once across all training trajectories becomes a node.
This yields $|\mathcal{V}| = 4{,}988$ nodes from the $\sim$16,000-tool library, reflecting the subset of tools actually used by successful agents.

\textbf{Directed edges.}
For each consecutive pair $(t_{\pi(i)}, t_{\pi(i+1)})$ in a trajectory, we add a directed edge $e = (t_{\pi(i)} \to t_{\pi(i+1)})$ representing the empirical observation that tool $t_{\pi(i)}$ immediately preceded $t_{\pi(i+1)}$ in a successful execution.
The raw co-occurrence count is:
\begin{equation}
    c(t_a, t_b) = \sum_{i=1}^{M} \sum_{j=1}^{L^{(i)}-1}
    \mathbf{1}\!\left[t_{\pi(j)}^{(i)} = t_a \wedge t_{\pi(j+1)}^{(i)} = t_b\right].
    \label{eq:cocount}
\end{equation}

\textbf{Edge weights.}
Raw co-occurrence counts are normalized into conditional transition probabilities to correct for tool popularity bias:
\begin{equation}
    w(t_a, t_b) = P(t_b \mid t_a) = \frac{c(t_a,\, t_b)}{\sum_{t' \in \mathcal{V}} c(t_a,\, t')}.
    \label{eq:transweight}
\end{equation}
$w(t_a, t_b)$ is the empirical probability that $t_b$ immediately follows $t_a$ in a successful workflow.
Self-loops are excluded.
We do not claim that every edge proves argument-level data flow; immediate transitions are a scalable proxy for workflow precedence.

The resulting graph $\mathcal{G}$ has $|\mathcal{V}| = 4{,}988$ nodes and $|\mathcal{E}| = 39{,}034$ directed edges.
The full construction procedure is summarized in Algorithm~\ref{alg:skillgraph}.

\begin{algorithm}[t]
\caption{SkillGraph Construction}
\label{alg:skillgraph}
\begin{algorithmic}
\STATE \textbf{Input:} Successful trajectories $\{(q^{(i)}, \tau^{(i)})\}_{i=1}^{M}$
\STATE \textbf{Output:} Directed weighted graph $\mathcal{G} = (\mathcal{V}, \mathcal{E}, \mathbf{W})$
\STATE Initialize count matrix $c(\cdot,\cdot) \leftarrow 0$
\FOR{each trajectory $\tau^{(i)} = [t_1, \ldots, t_{L^{(i)}}]$}
    \STATE Deduplicate while preserving order
    \FOR{$j = 1$ to $L^{(i)}-1$}
        \STATE $c(t_j,\, t_{j+1}) \mathrel{+}= 1$
    \ENDFOR
\ENDFOR
\STATE $\mathcal{V} \leftarrow \{t : \exists\, t' \text{ s.t. } c(t,t') > 0 \text{ or } c(t',t) > 0\}$
\STATE $\mathcal{E} \leftarrow \{(t_a, t_b) : c(t_a, t_b) > 0,\; t_a \neq t_b\}$
\STATE $w(t_a, t_b) \leftarrow c(t_a, t_b)\, /\, \textstyle\sum_{t'} c(t_a, t')$ for all $(t_a,t_b) \in \mathcal{E}$
\STATE \textbf{return} $\mathcal{G}$
\end{algorithmic}
\end{algorithm}

\subsection{Community Structure Analysis}

We verify that SkillGraph captures meaningful workflow structure rather than noisy co-occurrence patterns.
We apply the Louvain algorithm~\cite{louvain} to the undirected projection of $\mathcal{G}$ (edge weights averaged across both directions) to discover tool communities, and validate them against the API category labels provided by ToolBench~\cite{toolllm}.

\textbf{Structural results.}
Louvain discovers 216 communities.
Table~\ref{tab:graph_stats} summarizes the graph and community statistics.
The modularity $Q = 0.892$ far exceeds typical values for real-world social or web graphs ($Q \in [0.3, 0.7]$)~\cite{louvain}: tools form tight workflow groups with few cross-group transitions.

\begin{table}[!t]
\caption{SkillGraph Statistics}
\label{tab:graph_stats}
\centering
\small
\begin{tabular}{lc}
\toprule
\textbf{Property} & \textbf{Value} \\
\midrule
Nodes (tools)            & 4,988 \\
Directed edges           & 39,034 \\
Communities (Louvain)    & 216 \\
Modularity $Q$           & 0.892 \\
Mean community purity    & 0.779 \\
NMI (community vs.\ API category) & 0.695 \\
\bottomrule
\end{tabular}
\end{table}

\begin{figure}[!t]
\centering
\includegraphics[width=\columnwidth]{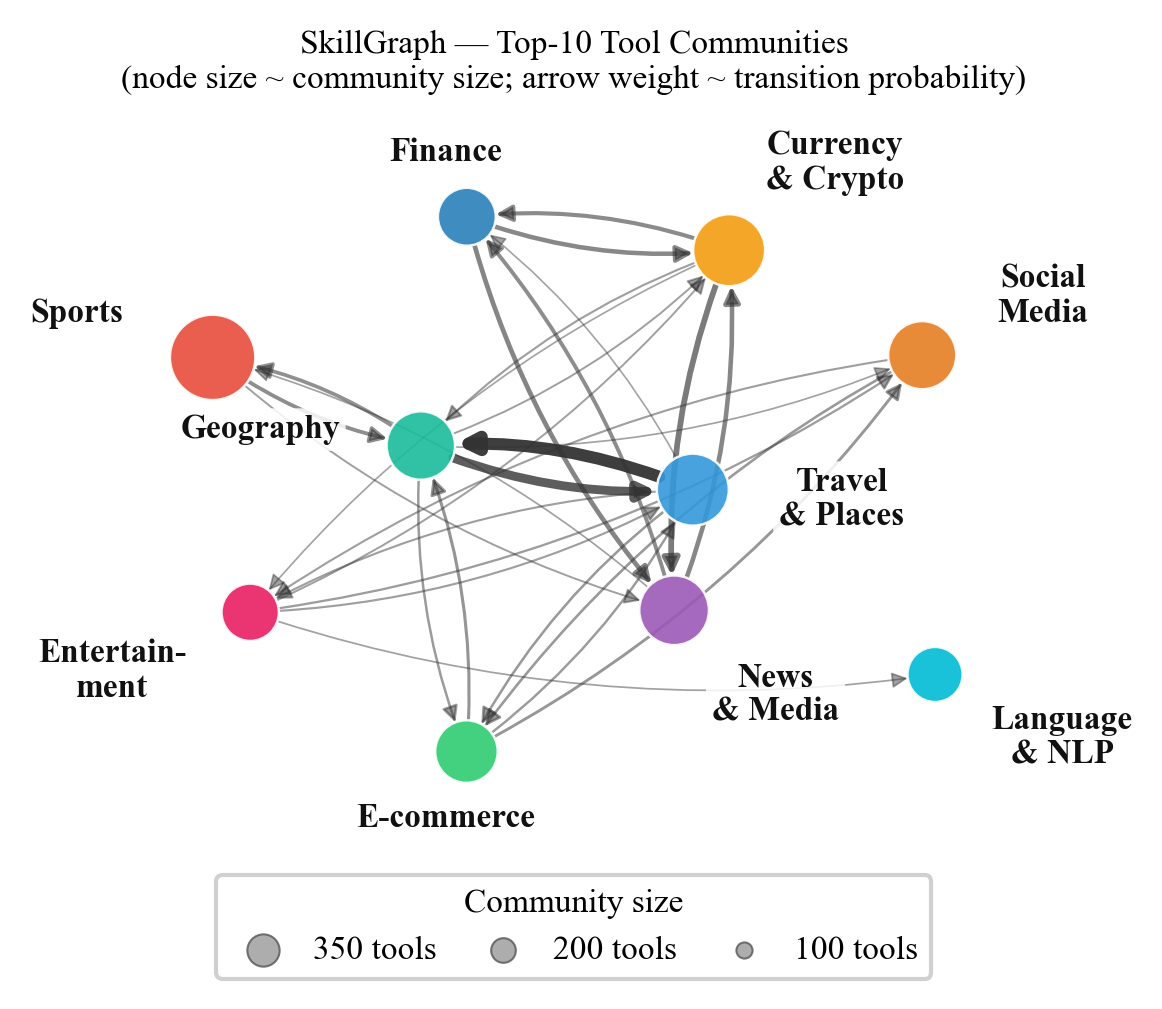}
\caption{Top-10 SkillGraph communities (out of 216 total; Louvain modularity $Q{=}0.892$).
Node size is proportional to community size (number of tools);
arrow weight reflects inter-community transition probability.
Communities align with API categories (NMI\,=\,0.695 vs.\ RapidAPI taxonomy):
domain-specific tools co-cluster naturally (e.g.\ \emph{Finance}
and \emph{Currency \& Crypto} exchange strongly, reflecting
multi-step trading workflows), validating that SkillGraph encodes
semantically coherent execution dependencies.}
\label{fig:community}
\end{figure}

\textbf{Semantic validity.}
For each community $C_k$, \emph{community purity} is the fraction of tools belonging to the dominant API category:
\begin{equation}
    \text{Purity}(C_k) = \frac{1}{|C_k|} \max_{c}\, |\{t \in C_k : \text{label}(t) = c\}|.
    \label{eq:purity}
\end{equation}
A mean purity of 0.779 and NMI of 0.695 confirm that SkillGraph communities align strongly with human-defined API categories---the graph captures semantically coherent workflow clusters, not arbitrary co-occurrence noise.
Figure~\ref{fig:community} visualises the top-10 communities and their inter-community transition structure.

\begin{figure}[!t]
\centering
\includegraphics[width=\columnwidth]{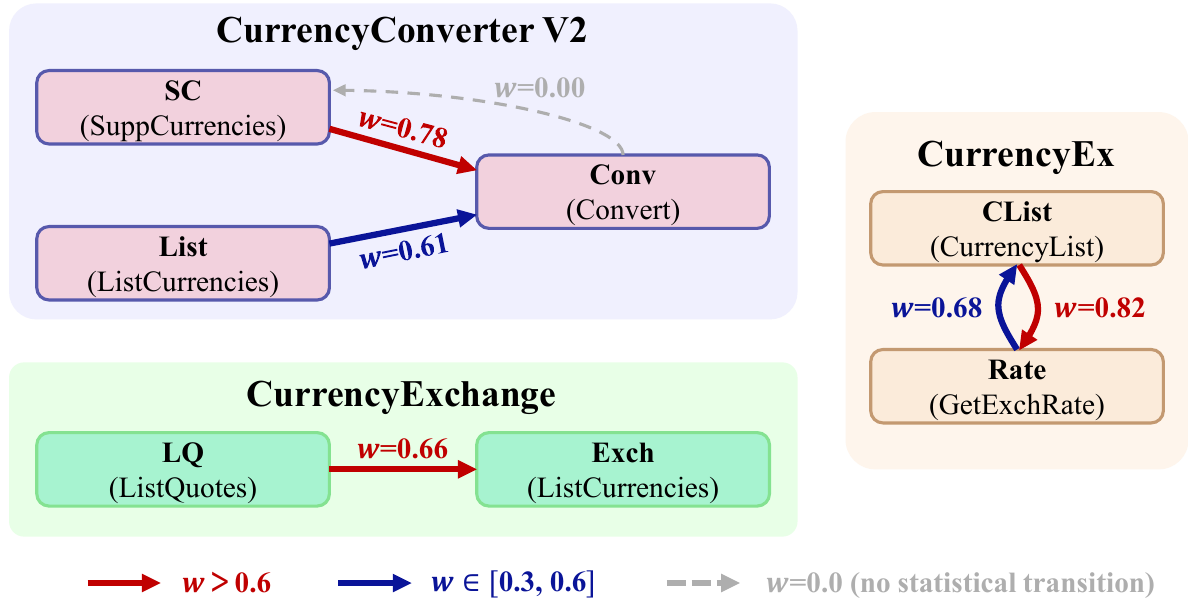}
\caption{SkillGraph subgraph for the currency-domain community (three API clusters).
Intra-cluster edges carry high statistical transition probability:
\texttt{SC}\,$\to$\,\texttt{Conv} ($w{=}0.78$, $n{=}23$) and
\texttt{CList}\,$\to$\,\texttt{Rate} ($w{=}0.82$) encode workflow dependencies
invisible in tool descriptions; clusters are self-contained with no
inter-cluster statistical co-occurrence.
The reverse edge \texttt{Conv}\,$\to$\,\texttt{SC} has $\text{transition\_prob}{=}0.00$
(dashed), confirming that SkillGraph captures execution \emph{order}, not merely
semantic similarity between tools.}
\label{fig:subgraph}
\end{figure}

\textbf{Case study.}
The workflow precedence relation identified in our motivating example (Fig.~\ref{fig:motivation}) is directly reflected in SkillGraph: the edge $(\texttt{SC} \to \texttt{Conv})$---abbreviating \texttt{supported\_currencies\_for\_currency\_converter\_v2} $\to$ \texttt{convert\_for\_currency\_converter\_v2}---has high statistical transition probability $w = P(\texttt{Conv}\mid\texttt{SC}) = 0.78$ ($n=23$), while the reverse edge has $w = P(\texttt{SC}\mid\texttt{Conv}) = 0.00$.
Both tools belong to the same currency-converter API community, confirming that the graph captures not only tool co-membership but also directed workflow order within a domain.

\subsection{Complementarity with Semantic Embeddings}

SkillGraph adds value only if it captures information not already present in semantic embeddings.
To verify this, we measure the Spearman rank correlation between the transition probability $w(t_a, t_b)$ of connected tool pairs and their semantic cosine similarity $\text{sim}(t_a, t_b) = \mathbf{e}_a^\top \mathbf{e}_b / (\|\mathbf{e}_a\|\|\mathbf{e}_b\|)$, where $\mathbf{e}_i$ is the sentence embedding of tool $t_i$'s description.

\begin{figure}[!t]
\centering
\resizebox{\columnwidth}{!}{%
\begin{tikzpicture}
\begin{axis}[
  xlabel={Semantic cosine similarity $\mathrm{sim}(t_a,t_b)$},
  ylabel={Transition probability $w(t_a,t_b)$},
  xlabel style={font=\small},
  ylabel style={font=\small},
  tick label style={font=\footnotesize},
  xmin=0, xmax=1,
  ymin=0, ymax=1,
  width=7.8cm, height=6.8cm,
  colormap={heatmap}{
    color(0)=(white)
    color(0.5)=(blue!30!white)
    color(1.5)=(cyan!60!blue)
    color(2.5)=(green!60!black)
    color(3.5)=(orange)
    color(4.2)=(red!80!black)
  },
  colorbar,
  colorbar style={
    font=\scriptsize,
    ylabel={$\log_{10}(\mathrm{count}+1)$},
    ylabel style={font=\scriptsize},
    width=0.25cm,
  },
  point meta min=0,
  point meta max=4.1,
  grid=both,
  grid style={gray!20, line width=0.3pt},
  major grid style={gray!30, line width=0.4pt},
  xtick={0,0.2,0.4,0.6,0.8,1.0},
  ytick={0,0.2,0.4,0.6,0.8,1.0},
]

%% ── 2D heatmap: each cell is a 0.05×0.05 square ──────────────────────────────
\addplot[
  matrix plot*,
  mesh/cols=20,
  mesh/rows=20,
  point meta=\thisrow{logcount},
] table [x=x, y=y, meta=logcount] {figure/complementarity_data.dat};

%% ── Annotate the SC→Conv example (sim=0.68, w=0.78) ─────────────────────────
\addplot[
  only marks,
  mark=star,
  mark size=3.5pt,
  color=red!80!black,
  line width=1.2pt,
] coordinates {(0.68, 0.78)};
\node[font=\scriptsize, text=red!80!black, anchor=south west]
  at (axis cs:0.70, 0.80)
  {\texttt{SC}$\!\to\!$\texttt{Conv}};

%% ── Spearman annotation ──────────────────────────────────────────────────────
\node[font=\scriptsize, fill=white, fill opacity=0.85, text opacity=1,
      inner sep=2pt, rounded corners=2pt,
      anchor=north east]
  at (axis cs:0.98, 0.98)
  {Spearman $\rho\!=\!-0.15$, $p\!<\!0.001$};

\end{axis}
\end{tikzpicture}}%
\caption{Transition probability $w(t_a,t_b)$ vs.\ semantic cosine similarity
for all 39,034 SkillGraph edges (log-scale density).
Spearman $\rho = -0.15$ ($p < 0.001$): the two signals are slightly negatively
correlated, confirming they capture complementary aspects of tool relationships.
High-probability transitions (e.g., \texttt{SC}\,$\to$\,\texttt{Conv}:
$w=0.78$, $\mathrm{sim}=0.68$) can co-occur with moderate semantic similarity,
yet the weak negative correlation shows that in aggregate, strong co-occurrence
does not imply semantic overlap.
This motivates the hybrid design in both stages of the framework.}
\label{fig:complementarity}
\end{figure}
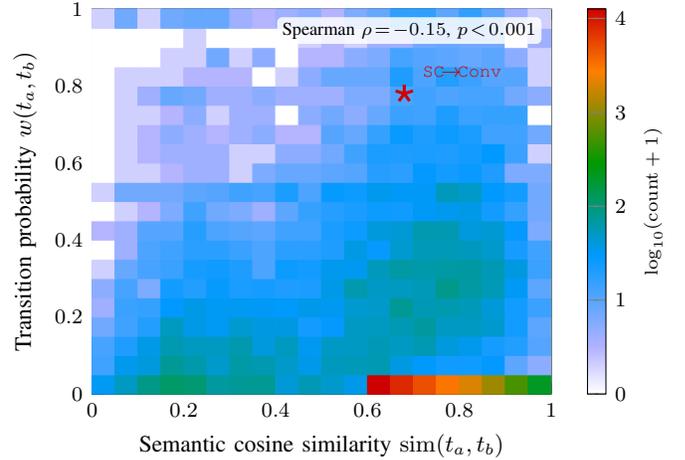

Across all $|\mathcal{E}| = 39{,}034$ edges, the Spearman correlation is $\rho = -0.15$ ($p < 0.001$), indicating a slight negative relationship: high transition probability does not predict high semantic similarity.
High-probability transitions can connect tools with varying degrees of semantic similarity (e.g., \texttt{SC} and \texttt{Conv}: $w = 0.78$, $\text{sim} = 0.68$), yet the weak negative aggregate correlation confirms that execution co-occurrence is not explained by semantic overlap.
SkillGraph and semantic embeddings thus carry complementary information, which motivates the hybrid design in both stages.
Figure~\ref{fig:complementarity} visualises the full joint distribution of transition probability and semantic similarity across all 39,034 edges, confirming that high-probability transitions populate all levels of semantic similarity---the two signals are genuinely complementary.

% ============================================================
\section{Two-Stage Decoupled Framework}
\label{sec:framework}
% ============================================================

\subsection{Motivation for Decoupling}

Single-stage methods cannot jointly optimize selection and ordering.
Semantic-only retrieval ranks tools by query relevance, achieving strong Set-F1 but ordering them incorrectly (Kendall-$\tau = 0.042$ on ToolBench; negative on API-Bank).
Beam search over SkillGraph transitions improves transition accuracy but drops selection quality: the beam follows edge weights, not query relevance, so it prunes tools that are relevant but weakly connected in the graph.

Selection and ordering pull in opposite directions in any shared scoring function: prioritizing query-relevance scores hurts ordering; prioritizing transition scores hurts coverage.
We decouple the two by giving each stage one job:
\begin{itemize}
    \item \textbf{Stage 1} maximizes $\text{Set-F1}(\hat{\mathcal{S}}, \mathcal{S}^*)$ using hybrid graph-semantic retrieval.
    \item \textbf{Stage 2} maximizes ordering quality \emph{on the fixed set} $\hat{\mathcal{S}}$ returned by Stage 1, using a learned pairwise reranker with SkillGraph features.
\end{itemize}
Fixing the candidate set after Stage~1 means Stage~2 can only reorder; it cannot change which tools are selected.

\begin{figure*}[!t]
\centering
\includegraphics[
  width=\textwidth,
  trim=106.49bp 641.42bp 309.46bp 491.44bp,
  clip
]{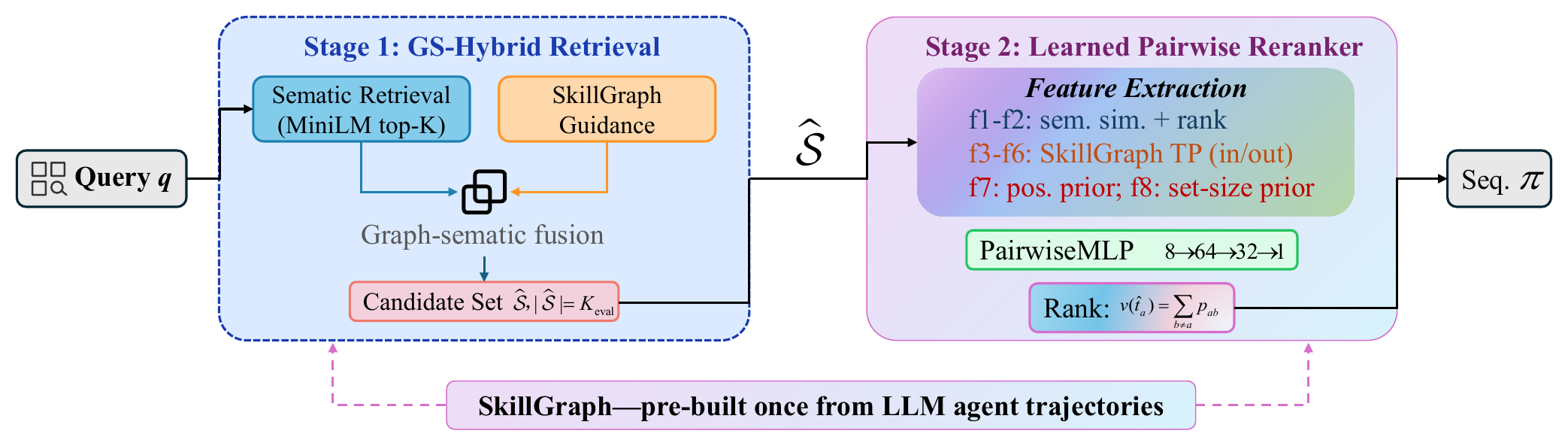}
\caption{Conceptual overview of the two-stage decoupled framework.
\textbf{Stage~1} (GS-Hybrid) constructs a candidate tool set $\hat{\mathcal{S}}$
of up to $K_{\mathrm{eval}}$ tools by combining dense semantic retrieval
with SkillGraph-guided candidate construction.
\textbf{Stage~2} (Learned Reranker) orders the candidates via a
PairwiseMLP that combines eight features: semantic similarity and rank
(f1--f2), SkillGraph transition probabilities (f3--f6), and positional
priors (f7--f8).
Decoupling the stages allows independent optimisation of set coverage
and sequence ordering, eliminating the single-stage selection-ordering
trade-off.}
\label{fig:framework}
\end{figure*}

Figure~\ref{fig:framework} provides a conceptual overview; the exact implementation details of both stages are given below.

\subsection{Stage 1: Graph-Semantic Hybrid Retrieval}

Stage~1 uses \textbf{Graph-Semantic Hybrid Retrieval (GS-Hybrid)} to construct the candidate set $\hat{\mathcal{S}}$, which is then fixed for Stage~2.

\textbf{Step 1: Semantic candidate pool.}
Each tool $t_i \in \mathcal{T}$ and the query $q$ are encoded with a pre-trained sentence encoder~\cite{sentencetransformers}:
\begin{equation}
    s_{\text{sem}}(q, t_i) = \frac{\mathbf{e}_q^\top \mathbf{e}_i}{\|\mathbf{e}_q\| \|\mathbf{e}_i\|},
    \label{eq:sem_score}
\end{equation}
where $\mathbf{e}_q, \mathbf{e}_i \in \mathbb{R}^d$ are the respective embeddings.
Rather than selecting only the final $K_{\mathrm{eval}}$ tools at this point, GS-Hybrid first retrieves a larger semantic pool
\[
\mathcal{C}_{\text{sem}} = \text{TopK}\!\left(q,\, K_{\text{pool}}\right), \qquad
K_{\text{pool}} = \max(K_{\mathrm{eval}} + 2,\, c \cdot K_{\mathrm{eval}}),
\]
where $c$ is a small multiplier.

\textbf{Step 2: SkillGraph-guided subgraph construction.}
We induce a SkillGraph subgraph over $\mathcal{C}_{\text{sem}}$.
If the induced subgraph has multiple weakly connected components, GS-Hybrid inserts a small number of bridge tools found along short graph paths between high-similarity nodes from adjacent components.
This enriches the candidate pool with tools that are weakly described semantically but important for connecting plausible execution chains.

\textbf{Step 3: Greedy provisional sequencing.}
GS-Hybrid builds a provisional sequence by starting from the highest-similarity tool and repeatedly appending the remaining tool with the largest hybrid score
\begin{equation}
    s_{\text{S1}}(t_j \mid t_i, q, p)
    = \alpha_{\text{S1}} \, w_{\text{loc}}(t_i, t_j)
    + (1 - \alpha_{\text{S1}})\, s_{\text{sem}}(q, t_j)
    + \gamma \, b_{\text{pos}}(t_j, p),
    \label{eq:stage1score}
\end{equation}
where $w_{\text{loc}}(t_i, t_j)$ is the local SkillGraph edge weight in the induced subgraph (using half weight for a reverse-only edge, as in the implementation), $p$ is the normalized output position, and $b_{\text{pos}}$ is a position bonus derived from training trajectories.
After truncation or padding to $K_{\mathrm{eval}}$, we discard the provisional order and pass only its tool set $\hat{\mathcal{S}}$ to Stage~2. Stage~1 is therefore evaluated with set metrics, while Stage~2 handles the final ordering metrics.

\subsection{Stage 2: Learned Pairwise Reranker}
\label{sec:framework-lr}

Stage~2 orders the fixed candidate set $\hat{\mathcal{S}} = \{\hat{t}_1, \ldots, \hat{t}_K\}$ from Stage~1 by learning pairwise preferences from SkillGraph and semantic features.

\textbf{Problem reduction.}
We frame ordering as pairwise classification: a binary classifier predicts, for each pair $(\hat{t}_a, \hat{t}_b)$, which tool runs first.
The final order is determined by sorting tools by their aggregated pairwise scores~\cite{ranknet}.

\textbf{Feature construction.}
For each tool $\hat{t}_a \in \hat{\mathcal{S}}$, we extract a per-tool feature vector $\mathbf{f}_a \in \mathbb{R}^8$:

\begin{enumerate}
    \item $s_{\text{sem}}(q, \hat{t}_a)$ — cosine similarity to query $q$
    \item $\text{rank}_{\text{sem}}(\hat{t}_a) / (K{-}1)$ — normalised retrieval rank ($0 =$ highest)
    \item $\textstyle\sum_{t' \in \hat{\mathcal{S}} \setminus \{\hat{t}_a\}} w(\hat{t}_a, t')$ — total outgoing SkillGraph weight to candidates
    \item $\textstyle\sum_{t' \in \hat{\mathcal{S}} \setminus \{\hat{t}_a\}} w(t', \hat{t}_a)$ — total incoming SkillGraph weight from candidates
    \item $\max_{t' \in \hat{\mathcal{S}} \setminus \{\hat{t}_a\}} w(\hat{t}_a, t')$ — strongest outgoing transition to a candidate
    \item $\max_{t' \in \hat{\mathcal{S}} \setminus \{\hat{t}_a\}} w(t', \hat{t}_a)$ — strongest incoming transition from a candidate
    \item $\bar{p}(\hat{t}_a)$ — mean normalised position in training trajectories ($0 =$ always first)
    \item $|\hat{\mathcal{S}}| / 10$ — normalised candidate set size (context signal)
\end{enumerate}
Features 1--2 measure query-tool relevance; features 3--6 reflect SkillGraph transition context within the candidate set; features 7--8 supply positional and set-size priors.
The pairwise input for pair $(\hat{t}_a, \hat{t}_b)$ is the antisymmetric difference $\mathbf{f}_{ab} = \mathbf{f}_a - \mathbf{f}_b \in \mathbb{R}^8$, ensuring $p_{ab} = 1 - p_{ba}$ by construction.

\textbf{Model architecture.}
The Learned Reranker (LR) is a 3-layer MLP:
\begin{equation}
    p_{ab} = \sigma\!\left(\text{MLP}(\mathbf{f}_{ab})\right), \quad
    \text{MLP}: \mathbb{R}^8 \to \mathbb{R}^{64} \to \mathbb{R}^{32} \to \mathbb{R},
    \label{eq:mlp}
\end{equation}
where $\sigma$ is the sigmoid function and $p_{ab} \in (0,1)$ is the predicted probability that $\hat{t}_a$ precedes $\hat{t}_b$.
ReLU activations are applied after each hidden layer.

\textbf{Training.}
For each training trajectory $\tau^{(i)}$, we generate all ordered pairs $(t_a, t_b)$ with ground-truth label $y_{ab} = \mathbf{1}[r_{S^*}(t_a) < r_{S^*}(t_b)]$, and minimize binary cross-entropy:
\begin{equation}
    \mathcal{L} = -\frac{1}{|\mathcal{P}|} \sum_{(a,b) \in \mathcal{P}} \left[ y_{ab} \log p_{ab} + (1 - y_{ab}) \log(1 - p_{ab}) \right],
    \label{eq:loss}
\end{equation}
where $\mathcal{P}$ is the set of all generated pairs across training trajectories.

\textbf{Inference.}
For a test query, we compute $p_{ab}$ for all $\binom{K}{2}$ pairs in $\hat{\mathcal{S}}$, assign each tool an aggregate score $v(\hat{t}_a) = \sum_{b \neq a} p_{ab}$, and sort tools in descending order of $v$.
This costs $O(K^2)$ time---negligible since $K \leq 10$ in practice.

\subsection{Alternative Stage-2 Methods}

We compare LR against four alternative Stage-2 ordering strategies, all operating on the same $\hat{\mathcal{S}}$ from Stage~1:

\textbf{Sem-Sort} orders tools by their semantic score $s_{\text{sem}}(q, \hat{t}_i)$ in descending order. This is equivalent to the pure semantic baseline when applied to GS-Hybrid's output.

\textbf{Hybrid-Rerank (HR)} is a non-learned Stage-2 reranker that scores each permutation $\pi$ of the fixed candidate set by
\begin{equation}
    s_{\text{HR}}(\pi)
    = \alpha_{\text{HR}} \sum_{i=1}^{K-1} w(\pi_i, \pi_{i+1})
    + (1 - \alpha_{\text{HR}}) \sum_{i=1}^{K} \frac{s_{\text{sem}}(q, \pi_i)}{i},
    \label{eq:hrscore}
\end{equation}
and returns the highest-scoring permutation (or a greedy approximation when $K$ is large).

\textbf{Optimal-Permutation (Opt-Perm)} exhaustively searches all $K!$ permutations of $\hat{\mathcal{S}}$ and selects the one maximizing the cumulative log transition score from SkillGraph. It is a graph-only exhaustive search baseline over the fixed candidate set, not an oracle using $S^*$.

\textbf{LLM-Reranker} prompts LLaMA-3.1-8B~\cite{llama3} with the query and candidate tool list (zero-shot or 3-shot) to output a ranked ordering. This tests whether parametric LLM knowledge can substitute for SkillGraph dependency information.

Unlike the other methods, LR learns when graph dependency should override semantic relevance, and vice versa---the alternatives either fix this trade-off by hand or delegate it to an LLM.

% ============================================================
\section{Experiments}
\label{sec:experiments}
% ============================================================

\subsection{Datasets}

\textbf{ToolBench.}
ToolBench~\cite{toolllm} is a large-scale tool-use benchmark constructed from 16,464 real-world APIs spanning 49 categories on RapidAPI.
We use the subset of successful DFSDT trajectories, split into 49,831 training and 9,965 test instances.
Each instance pairs a natural-language instruction with an ordered sequence of tool invocations executed by a GPT-3.5-based agent.
Ground-truth tool sequences range in length from 1 to 12 tools (mean 2.8).
ToolBench tests recommendation at scale: the tool library is large ($N \approx 16{,}000$), queries are diverse, and tool descriptions are noisy real-world API documentation.

\textbf{API-Bank Level-3.}
API-Bank~\cite{apibank} is a benchmark of tool-augmented LLM tasks at three difficulty levels.
We use Level-3, which contains 50 multi-step instructions requiring \emph{sequential} tool invocations with explicit inter-tool data dependencies (e.g., the output of one API is a required input to the next).
The tool library has 21 tools across productivity, calendar, and communication domains.
We evaluate using leave-one-out cross-validation (LOO-CV): for each held-out instance, graph-based transition statistics are estimated from the remaining 49 Level-3 trajectories.
The Learned Reranker transfers its weights and positional priors from ToolBench without API-Bank fine-tuning, making API-Bank a low-resource transfer test on a different tool set and domain rather than a pure zero-resource reuse of an unchanged ToolBench graph.

\subsection{Baselines}

We compare against the following methods, organized by architecture type.

\textbf{Single-stage methods} predict the full ordered sequence in one pass:
\begin{itemize}
    \item \textbf{Semantic Only}: retrieves top-$K$ tools by cosine similarity (Eq.~\eqref{eq:sem_score}) and sorts by score.
    \item \textbf{BM25}: replaces dense embeddings with sparse BM25 retrieval over tool descriptions~\cite{bm25review}; ordering follows retrieval score.
    \item \textbf{Beam Search}: performs beam search over SkillGraph transition edges, starting from the top-1 semantic match and expanding greedily by transition weight. Combines retrieval and ordering in a single pass.
    \item \textbf{Hybrid Sem-Graph}: runs the single-stage GS-Hybrid planner end-to-end and returns its provisional sequence directly, without a separate Stage-2 reranker.
\end{itemize}

\textbf{Two-stage methods} all use GS-Hybrid (Section~\ref{sec:framework}-B) as Stage~1, varying only Stage~2:
\begin{itemize}
    \item \textbf{GS-Hybrid + Sem-Sort}: orders the Stage-1 candidate set by $s_{\text{sem}}$. Equivalent to Semantic Only on GS-Hybrid's output.
    \item \textbf{GS-Hybrid + Hyb-Rerank (HR)}: applies the non-learned Stage-2 reranker defined in Eq.~\eqref{eq:hrscore} to the fixed GS-Hybrid candidate set.
    \item \textbf{GS-Hybrid + Opt-Perm}: exhaustively evaluates all $K!$ permutations and selects the one with the largest cumulative log transition score under the graph. This isolates the best fixed graph-only ordering over the Stage-1 candidate set.
    \item \textbf{GS-Hybrid + LR (ours)}: our Learned Reranker (Section~\ref{sec:framework-lr}).
\end{itemize}

\textbf{Extended baseline} probes an alternative Stage-2 strategy:
\begin{itemize}
    \item \textbf{GS-Hybrid + LLaMA (0-shot / 3-shot)}: replaces Stage~2 with LLaMA-3.1-8B~\cite{llama3} prompted to rank the candidate tools. Evaluated on a 300-instance sample of ToolBench for computational feasibility.
\end{itemize}

\subsection{Implementation Details}

\textbf{Embeddings.}
All tools and queries are encoded with \texttt{all-MiniLM-L6-v2}~\cite{sentencetransformers} ($d = 384$), producing $L_2$-normalized embeddings. Tool embeddings are pre-computed and cached.

\textbf{Stage-1 hyperparameters.}
For GS-Hybrid, we use the tuned graph-search parameters $c = 3$, $\alpha_{\text{S1}} = 0.5$, and $\gamma = 0.1$, so the semantic candidate pool size is $K_{\text{pool}} = \max(K_{\mathrm{eval}} + 2, 3K_{\mathrm{eval}})$.
The Stage-2 Hybrid-Rerank coefficient is tuned separately and set to $\alpha_{\text{HR}} = 0.4$ (Section~\ref{sec:analysis}).

\textbf{Learned Reranker training.}
The PairwiseMLP is trained for up to 30 epochs with early stopping (patience~$= 5$) using Adam (learning rate $10^{-3}$, batch size 2048).
Training pairs are generated from all ordered tool pairs in each ground-truth trajectory; for every positive pair $(t_a, t_b)$ with $t_a$ preceding $t_b$, we also include the reversed pair $(t_b, t_a)$ as a negative example.
Training takes under 5 minutes on a single GPU.

\textbf{API-Bank protocol details.}
For each LOO fold on API-Bank, tool descriptions are extracted from API-Bank metadata and fall back to the tool name when a description is missing.
Graph-based features on the held-out instance are computed only from the other 49 Level-3 trajectories in that fold.
The LR model is not retrained on API-Bank: it reuses the ToolBench-trained checkpoint and ToolBench-derived position statistics zero-shot; because API-Bank tools are largely unseen in ToolBench, these positional priors rarely fire and the transfer relies mainly on semantic and within-fold transition features.

\textbf{LLM baselines.}
LLaMA-3.1-8B is run in 4-bit quantization (bfloat16) on a single A100.
The zero-shot prompt presents the query and candidate tool names and asks for a ranked list; the 3-shot prompt additionally includes three in-context examples drawn from the training set.

\textbf{Statistical testing.}
All pairwise method comparisons use bootstrap resampling with 10,000 iterations (two-sided).
We report $p$-values and flag significance at $p < 0.05$ ($^*$) and $p < 0.01$ ($^{**}$).

\subsection{Main Results on ToolBench}

\begin{table*}[!t]
\caption{Tool Sequence Recommendation Results on ToolBench (9,965 Test Instances).
Best results in \textbf{bold}.
Significance vs.\ GS-Hybrid+Sem-Sort (bootstrap, 10K resamples):
$^{*}$~$p{<}0.05$, $^{**}$~$p{<}0.01$.}
\label{tab:main}
\centering
\small
\begin{tabular}{lccccc}
\toprule
\textbf{Method} & \textbf{Set-F1} & \textbf{Ord.Prec} & \textbf{Kendall-$\tau$} & \textbf{Trans.Acc} & \textbf{1st.Acc} \\
\midrule
\multicolumn{6}{l}{\textit{Single-Stage Methods}} \\
Semantic Only         & 0.266 & 0.123 & 0.042 & 0.070 & 0.262 \\
BM25                  & 0.215 & 0.073 & $-$0.001 & 0.041 & 0.143 \\
Beam Search           & 0.240 & 0.149 & 0.089 & 0.099 & 0.206 \\
Hybrid Sem-Graph      & 0.271 & 0.130 & 0.044 & 0.082 & 0.262 \\
\midrule
\multicolumn{6}{l}{\textit{Two-Stage Methods (Stage~1: GS-Hybrid)}} \\
GS-Hybrid + Sem-Sort              & 0.271 & 0.133 & 0.051 & 0.083 & 0.262 \\
GS-Hybrid + Hyb-Rerank            & 0.271 & 0.152$^{**}$ & 0.089$^{**}$ & 0.097$^{**}$ & 0.252 \\
GS-Hybrid + Opt-Perm              & 0.271 & 0.151 & 0.086 & 0.097 & 0.227 \\
GS-Hybrid + LR (ours)             & \textbf{0.271} & \textbf{0.156}$^{**}$ & \textbf{0.096}$^{**}$ & \textbf{0.098}$^{**}$ & 0.234 \\
\bottomrule
\end{tabular}
\end{table*}

Table~\ref{tab:main} shows that GS-Hybrid~+~LR is the sole Pareto-optimal method on ToolBench.

\textbf{Value of Stage~2.}
Against GS-Hybrid~+~Sem-Sort, which uses the same Stage-1 candidate set, LR improves Ord.Prec from 0.133 to 0.156 and Kendall-$\tau$ from 0.051 to 0.096; both gains are significant ($p < 0.01$, bootstrap).
The improvement therefore comes entirely from Stage-2 ordering.

\textbf{Single-stage trade-off.}
Beam Search illustrates the joint selection-ordering trade-off: its Kendall-$\tau$ (0.089) approaches LR but at the cost of lower Set-F1 (0.240 vs.\ 0.271), a 12\% regression.
Hybrid Sem-Graph improves Set-F1 via SkillGraph-guided candidate construction but its ordering quality (Kendall-$\tau = 0.044$) barely exceeds Semantic Only (0.042), confirming that a mixed score applied uniformly to all tools cannot resolve the signal gap.

\textbf{LR vs.\ HR.}
LR improves over the non-learned HR baseline on Ord.Prec ($p = 0.025$) and Kendall-$\tau$ ($p = 0.024$): \emph{learning} the interaction between semantic and graph features outperforms the fixed linear combination.

\textbf{1st.Acc trade-off.}
LR shows a regression on First-Tool Accuracy (0.234 vs.\ 0.262 for Sem-Sort).
Because the pairwise objective optimizes \emph{global} rank agreement, it may deprioritize the first position.
We analyze this trade-off in Section~\ref{sec:analysis}.

\subsection{Low-Resource Validation on API-Bank}

\begin{table*}[!t]
\caption{Tool Sequence Recommendation Results on API-Bank Level-3 (50 instances, LOO-CV). In each fold, graph-based transition statistics are estimated from the other 49 API-Bank instances, while GS-Hybrid+LR reuses ToolBench-trained reranker weights zero-shot.}
\label{tab:apibank}
\centering
\small
\begin{threeparttable}
\begin{tabular}{lcccc}
\toprule
\textbf{Method} & \textbf{Set-F1} & \textbf{Ord.Prec} & \textbf{Kendall-$\tau$} & \textbf{Trans.Acc} \\
\midrule
BM25                   & 0.590 & 0.100 & $-$0.100 & 0.10 \\
Semantic Only          & 0.793 & 0.133 & $-$0.433 & 0.15 \\
Hybrid Sem-Graph       & 0.803 & 0.193 & $-$0.293 & 0.19 \\
GS-Hybrid + Hyb-Rerank & 0.803 & 0.173 & $-$0.333 & 0.19 \\
GS-Hybrid + LR (ours)  & \textbf{0.803} & \textbf{0.647} & \textbf{+0.613} & \textbf{0.62} \\
\midrule
LLaMA-3.1-8B (3-shot)  & 0.945\tnote{$\ddagger$} & 0.747 & -- & -- \\
\bottomrule
\end{tabular}
\begin{tablenotes}
\footnotesize
\item[] Bootstrap 95\% CI for Kendall-$\tau$: Semantic Only $[-0.607,-0.247]$, Hybrid Sem-Graph $[-0.480,-0.107]$, GS-Hybrid + HR $[-0.507,-0.147]$, GS-Hybrid + LR $[0.487,0.740]$.
\item[$\ddagger$] LLaMA uses oracle candidate set (ground-truth tools), inflating Set-F1.
\end{tablenotes}
\end{threeparttable}
\end{table*}

Table~\ref{tab:apibank} shows the selection-ordering signal gap in its strongest form.
On API-Bank Level-3, all semantic-based methods, including Hybrid Sem-Graph, yield negative Kendall-$\tau$, indicating systematic order inversion under strong inter-tool dependencies. Their 95\% bootstrap intervals also remain strictly negative, whereas GS-Hybrid+LR remains strictly positive.
GS-Hybrid~+~LR reverses this pattern: Kendall-$\tau$ rises to $+0.613$, a 0.946 absolute improvement over HR, while Ord.Prec and Trans.Acc improve by factors of 3.7$\times$ and 3.3$\times$.
SkillGraph's workflow-precedence prior, combined with learned pairwise ranking, captures structural information that embedding similarity cannot.

Regarding LLaMA-3.1-8B (3-shot): its Set-F1 of 0.945 is not comparable to other methods because it receives the \emph{oracle} ground-truth tool set as input, bypassing Stage~1 entirely.
This provides an approximate upper bound on Set-F1 for this benchmark.
Even with this advantage, LR achieves competitive ordering (Ord.Prec 0.647 vs.\ 0.747 for LLaMA), while operating without oracle tool access.
We caution that the 50-instance size of API-Bank limits statistical power; these results are best interpreted as evidence of qualitative behavior rather than precise quantitative comparison.
This setting should also be interpreted carefully: the graph features are estimated within each LOO fold from API-Bank itself, whereas only the LR weights transfer zero-shot from ToolBench.

\subsection{LLM Stage-2 Reranker Comparison}

\begin{table*}[!t]
\caption{Comparison with LLM-Based Stage-2 Reranking on ToolBench (Sampled $n=300$, Fair: All Methods Use Identical Stage-1 Output). Set-F1 is equal across LR/HR/LLaMA-0shot (same Stage-1); LLaMA-3shot differs marginally due to list-reordering affecting tool parsing.}
\label{tab:llm}
\centering
\small
\begin{tabular}{lcccc}
\toprule
\textbf{Method} & \textbf{Set-F1} & \textbf{Ord.Prec} & \textbf{Kendall-$\tau$} & \textbf{Trans.Acc} \\
\midrule
GS-Hybrid + LR (ours)       & 0.254 & \textbf{0.140} & \textbf{0.097} & \textbf{0.084} \\
GS-Hybrid + Hyb-Rerank      & 0.254 & 0.132 & 0.081 & 0.082 \\
GS-Hybrid + LLaMA (0-shot)  & 0.254 & 0.116 & 0.048 & 0.083 \\
GS-Hybrid + LLaMA (3-shot)  & 0.257 & 0.110 & 0.043 & 0.071 \\
\bottomrule
\end{tabular}
\end{table*}

Table~\ref{tab:llm} compares Stage-2 ordering strategies under a strictly controlled setup: all methods receive identical Stage-1 output, so Set-F1 is equal by construction and differences reflect purely ordering quality.

LR outperforms both LLaMA variants on every ordering metric.
LLaMA-3.1-8B (3-shot) \emph{performs worse than zero-shot} (Kendall-$\tau$: 0.043 vs.\ 0.048), and \emph{both LLaMA variants fall below the non-learned HR baseline} (Kendall-$\tau$: 0.081).
In-context examples do not help the model generalize tool execution order---few-shot demonstrations cannot convey the full dependency structure in SkillGraph.

LLMs reason about tool order from \emph{semantic plausibility}---which tool ``sounds like'' it should come first---whereas correct tool order is determined by \emph{data-flow dependencies}: which tool's output the next tool needs.
SkillGraph makes these precedence cues explicit through empirical transition statistics, enabling LR to learn order directly from execution outcomes rather than inferring it from descriptions.
A lightweight 3-layer MLP trained on 8 SkillGraph-derived features outperforms an 8-billion-parameter language model for this task.

\subsection{Ablation Study}

\begin{table}[!t]
\caption{Ablation Study on ToolBench (9,965 test instances).
\textit{Top}: Stage-2 Hybrid-Rerank mixing coefficient $\alpha$ (Stage-1: GS-Hybrid fixed).
Set-F1 improves from 0.266 ($\alpha{=}0$) to 0.271 ($\alpha{=}0.4$); ordering metrics
are $\alpha$-stable for $\alpha \in [0.1, 1.0]$, confirming stage decoupling.
\textit{Middle}: Stage-2 strategy comparison (Stage-1: GS-Hybrid, $\alpha{=}0.4$).
\textit{Bottom}: LR feature-group ablation (inference-time zeroing; Set-F1 fixed at 0.271 for all).}
\label{tab:ablation}
\centering
\small
\setlength{\tabcolsep}{4pt}%
\begin{tabular}{l@{\hspace{5pt}}c@{\hspace{5pt}}c@{\hspace{5pt}}c}
\toprule
\textbf{Configuration} & \textbf{Ord.Prec} & \textbf{Kendall-$\tau$} & \textbf{Trans.Acc} \\
\midrule
\multicolumn{4}{l}{\textit{Stage-2 HR $\alpha$ sensitivity (Stage-1: GS-Hybrid fixed)}} \\
$\alpha = 0.0$              & 0.134 & 0.052 & 0.084 \\
$\alpha = 0.1$              & 0.152 & 0.087 & 0.097 \\
$\alpha = 0.4$ \textbf{(ours)} & 0.154 & 0.092 & 0.098 \\
$\alpha = 0.7$              & 0.153 & 0.089 & 0.097 \\
$\alpha = 1.0$              & 0.152 & 0.087 & 0.097 \\
\midrule
\multicolumn{4}{l}{\textit{Stage-2 strategy ($\alpha{=}0.4$; Stage-1: GS-Hybrid)}} \\
Sem-Sort                        & 0.133 & 0.051 & 0.083 \\
Hybrid-Rerank (non-learned)     & 0.152 & 0.089 & 0.097 \\
Opt-Perm (graph-only exhaustive) & 0.151 & 0.086 & 0.097 \\
Learned Reranker \textbf{(ours)}& \textbf{0.156} & \textbf{0.096} & \textbf{0.098} \\
\midrule
\multicolumn{4}{l}{\textit{LR feature ablation (Set-F1 $= 0.271$ for all rows)}} \\
Full LR (all 8 features)            & \textbf{0.157} & \textbf{0.096} & \textbf{0.099} \\
$-$ Graph transitions (f3--f6)      & 0.149 & 0.081 & 0.095 \\
$-$ Positional priors (f7--f8)      & 0.152 & 0.086 & 0.095 \\
$-$ Semantic features (f1--f2)      & 0.154 & 0.090 & 0.099 \\
\bottomrule
\end{tabular}
\smallskip\par\noindent\footnotesize
$^\dagger$Results across blocks may differ by ${\leq}0.003$ due to run-to-run variation in Stage-1 retrieval (approximate nearest-neighbour search); this affects all methods including non-learned ones. LR additionally exhibits ${\leq}0.001$ variation from training stochasticity.
\end{table}

\begin{figure}[!t]
\centering
\resizebox{\columnwidth}{!}{%
\begin{tikzpicture}
\begin{axis}[
  xlabel={$\alpha$ (graph-transition weight in Hybrid-Rerank)},
  ylabel={Score},
  xlabel style={font=\small},
  ylabel style={font=\small},
  tick label style={font=\footnotesize},
  xmin=-0.05, xmax=1.05,
  ymin=0.040, ymax=0.172,
  xtick={0,0.2,0.4,0.6,0.8,1.0},
  ytick={0.06,0.08,0.10,0.12,0.14,0.16},
  xticklabel style={font=\footnotesize},
  yticklabel style={font=\footnotesize},
  width=7.8cm, height=6.0cm,
  %% Legend: vertical, bottom-right, below all three lines
  legend style={font=\scriptsize, at={(0.99,0.02)}, anchor=south east,
    row sep=0pt, inner sep=2pt, draw=gray!40},
  legend cell align=left,
  grid=both,
  grid style={gray!20, line width=0.4pt},
  major grid style={gray!35, line width=0.5pt},
  clip=false,
]

%% ── Shaded plateau (0.3–0.7) ────────────────────────────────────────────────
\addplot[draw=none, fill=gray!14, forget plot]
  coordinates {(0.3,0.040)(0.7,0.040)(0.7,0.172)(0.3,0.172)} \closedcycle;

%% Plateau label at the TOP of the shaded band — above all data lines
\node[font=\scriptsize, text=gray!60, anchor=south]
  at (axis cs:0.5, 0.12)
  {plateau $\alpha\!\in\![0.3,\,0.7]$};

%% ── Ordered Precision ────────────────────────────────────────────────────────
\addplot[color=red!75!black, mark=*, mark size=2.0pt,
         line width=1.4pt, mark options={fill=red!60}]
  coordinates {
    (0.0, 0.134)
    (0.1, 0.1515)
    (0.2, 0.1528)
    (0.3, 0.1535)
    (0.4, 0.1540)
    (0.5, 0.1532)
    (0.6, 0.1526)
    (0.7, 0.1527)
    (0.8, 0.1519)
    (0.9, 0.1514)
    (1.0, 0.1515)
  };
\addlegendentry{Ordered Precision}

%% ── Kendall-τ ────────────────────────────────────────────────────────────────
\addplot[color=orange!85!black, mark=square*, mark size=2.0pt,
         line width=1.4pt, dashed, mark options={fill=orange!55}]
  coordinates {
    (0.0, 0.0518)
    (0.1, 0.0869)
    (0.2, 0.0894)
    (0.3, 0.0909)
    (0.4, 0.0918)
    (0.5, 0.0903)
    (0.6, 0.0890)
    (0.7, 0.0892)
    (0.8, 0.0876)
    (0.9, 0.0866)
    (1.0, 0.0868)
  };
\addlegendentry{Kendall-$\tau$}

%% ── Transition Accuracy ──────────────────────────────────────────────────────
\addplot[color=green!60!black, mark=triangle*, mark size=2.2pt,
         line width=1.4pt, dotted, mark options={fill=green!45}]
  coordinates {
    (0.0, 0.0836)
    (0.1, 0.0967)
    (0.2, 0.0971)
    (0.3, 0.0977)
    (0.4, 0.0977)
    (0.5, 0.0973)
    (0.6, 0.0972)
    (0.7, 0.0973)
    (0.8, 0.0970)
    (0.9, 0.0969)
    (1.0, 0.0969)
  };
\addlegendentry{Trans.~Acc}

%% ── Best-α marker — label placed to the RIGHT of the dashed line ─────────────
\draw[red!65, dashed, line width=0.7pt]
  (axis cs:0.4,0.040) -- (axis cs:0.4,0.172);
\node[font=\scriptsize, text=red!70!black, anchor=south west]
  at (axis cs:0.41, 0.1545) {$\alpha^*\!=\!0.4$};

\end{axis}
\end{tikzpicture}}%
\caption{$\alpha$ sensitivity of GS-Hybrid+Hybrid-Rerank on ToolBench
(Stage~1 fixed; Stage~2 Hybrid-Rerank weight varied).
All three metrics jump sharply from $\alpha{=}0$ (semantic only)
to $\alpha{=}0.1$ (graph added), then plateau over
$\alpha\!\in\![0.3,\,0.7]$, confirming that Stage~1 and Stage~2
can be optimised independently.}
\label{fig:alpha}
\end{figure}
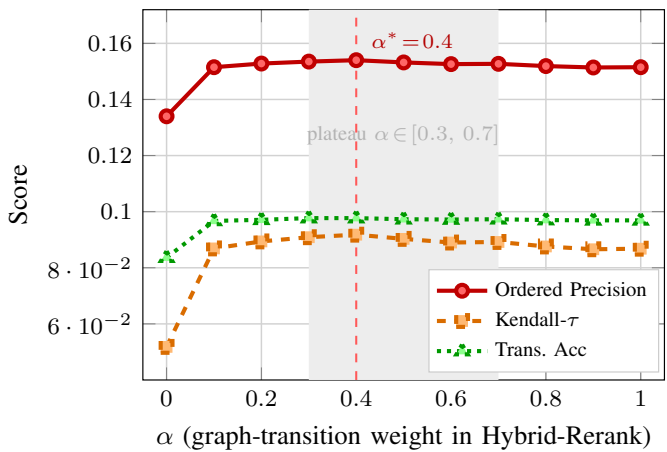

Table~\ref{tab:ablation} (top) isolates the effect of the Hybrid-Rerank mixing coefficient $\alpha$ while keeping GS-Hybrid Stage~1 fixed.
Ordering metrics are remarkably stable for $\alpha \in [0.1, 1.0]$ (Kendall-$\tau$ range: 0.087--0.092), collapsing only at $\alpha = 0.0$ (pure semantic, Kendall-$\tau = 0.052$) where graph guidance is removed from the reranker.
Once the GS-Hybrid candidate set is fixed, Stage-2 ordering quality is robust across a broad range of graph-semantic trade-offs.
Figure~\ref{fig:alpha} plots all three metrics as a function of $\alpha$, making the sharp jump at $\alpha{=}0{\to}0.1$ and the subsequent plateau visually clear.

Table~\ref{tab:ablation} (middle) compares Stage-2 strategies on the same GS-Hybrid Stage-1 output.
The progression Sem-Sort (0.051) $\to$ HR (0.089) $\to$ LR (0.096) confirms that each level of graph-awareness adds ordering quality, and LR exceeds the graph-only exhaustive Opt-Perm on Ord.Prec (0.156 vs.\ 0.151).

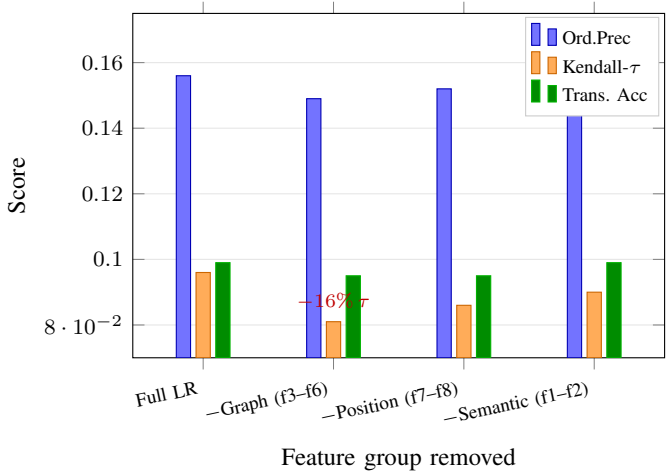
\begin{figure}[!t]
\centering
\resizebox{\columnwidth}{!}{%
\begin{tikzpicture}
\begin{axis}[
  ybar,
  bar width=5.2pt,
  width=8.4cm, height=6.0cm,
  xlabel={Feature group removed},
  ylabel={Score},
  xlabel style={font=\small},
  ylabel style={font=\small},
  tick label style={font=\footnotesize},
  symbolic x coords={Full LR, No Graph, No Position, No Semantic},
  xtick=data,
  xticklabels={Full LR, $-$Graph (f3--f6), $-$Position (f7--f8), $-$Semantic (f1--f2)},
  xticklabel style={font=\scriptsize, rotate=12, anchor=north east},
  ymin=0.070, ymax=0.175,
  ytick={0.08,0.10,0.12,0.14,0.16},
  yticklabel style={font=\footnotesize},
  legend style={font=\scriptsize, at={(0.99,0.99)}, anchor=north east,
    inner sep=2pt, row sep=-1pt, draw=gray!40},
  legend cell align=left,
  ymajorgrids=true,
  grid style={gray!20, line width=0.4pt},
  enlarge x limits=0.18,
]

\addplot[fill=blue!55, draw=blue!70!black]
  coordinates {
    (Full LR,     0.156)
    (No Graph,    0.149)
    (No Position, 0.152)
    (No Semantic, 0.154)
  };
\addlegendentry{Ord.Prec}

\addplot[fill=orange!65, draw=orange!80!black]
  coordinates {
    (Full LR,     0.096)
    (No Graph,    0.081)
    (No Position, 0.086)
    (No Semantic, 0.090)
  };
\addlegendentry{Kendall-$\tau$}

\addplot[fill=green!55!black, draw=green!70!black]
  coordinates {
    (Full LR,     0.099)
    (No Graph,    0.095)
    (No Position, 0.095)
    (No Semantic, 0.099)
  };
\addlegendentry{Trans.~Acc}

\node[font=\scriptsize, text=red!75!black, anchor=south]
  at (axis cs:No Graph, 0.081) {\raisebox{2pt}{$-16\%\,\tau$}};

\end{axis}
\end{tikzpicture}}%
\caption{LR feature-group ablation on ToolBench (Set-F1 fixed at 0.271 for all
variants; inference-time zeroing). Graph transition features (f3--f6) contribute
most: removing them drops Kendall-$\tau$ by $16\%$ and Ord.Prec by $5\%$.
Positional priors (f7--f8) rank second ($-11\%\,\tau$); semantic features (f1--f2)
contribute least ($-6\%\,\tau$), confirming SkillGraph as the primary ordering
signal.}
\label{fig:ablation_bar}
\end{figure}

Table~\ref{tab:ablation} (bottom) reports the LR feature-group ablation (zeroing each group at inference; Set-F1 remains 0.271 for all variants).
\textbf{Graph transition features} (f3--f6) contribute most: their removal drops Kendall-$\tau$ from 0.096 to 0.081 ($-16\%$) and Ord.Prec from 0.157 to 0.149 ($-5\%$), confirming SkillGraph as the primary ordering signal.
\textbf{Positional priors} (f7--f8) are second: Kendall-$\tau$ falls to 0.086 ($-11\%$), reflecting the value of empirical position statistics from training trajectories.
\textbf{Semantic features} (f1--f2) contribute least: Kendall-$\tau$ drops to 0.090 ($-6\%$), query relevance provides useful context but cannot determine order alone.
Figure~\ref{fig:ablation_bar} summarises the feature-group ablation visually.

% ============================================================
\section{Analysis}
\label{sec:analysis}
% ============================================================

\subsection{The Selection-Ordering Signal Gap}

We quantify the asymmetry between selection and ordering signals across all methods.
For Set-F1, the improvement from adding graph information is modest: Semantic Only (0.266) $\to$ GS-Hybrid (0.271), a relative gain of +1.9\%.
In contrast, the improvement on Kendall-$\tau$ from adding graph-guided reranking is substantial: Semantic Only (0.042) $\to$ GS-Hybrid+LR (0.096), a relative gain of +129\%.
On API-Bank, where workflow dependencies are explicit, the gap is even starker: Semantic Only achieves Kendall-$\tau = -0.433$, while GS-Hybrid+LR reaches $+0.613$---an absolute improvement of 1.046 from integrating SkillGraph.

Tool selection is a \emph{set membership} problem: query and relevant tools share domain vocabulary, making embedding similarity a natural proxy.
Tool ordering is a \emph{dependency resolution} problem: the correct order is determined by data-flow constraints (tool A produces data required by tool B), which are only weakly reflected in natural language descriptions.
SkillGraph's transition statistics capture these precedence cues where embedding similarity cannot.

\subsection{Sequence Length Analysis}

\begin{table*}[!t]
\caption{Kendall-$\tau$ by Ground-Truth Sequence Length on ToolBench.
GS-Hybrid+LR consistently improves over Semantic Only across all sequence lengths
by $119$--$134\%$.}
\label{tab:length}
\centering
\small
\begin{tabular}{lcccc}
\toprule
\textbf{Length bucket} & \textbf{Semantic Only} & \textbf{GS-Hybrid+HR} & \textbf{GS-Hybrid+LR} & \textbf{LR vs.\ Sem. (\%)} \\
\midrule
1--2 tools & 0.044 & 0.085 & 0.098 & $+$123\% \\
3--4 tools & 0.040 & 0.087 & 0.093 & $+$134\% \\
5$+$ tools & 0.049 & 0.111 & 0.108 & $+$119\% \\
\midrule
All lengths & 0.042 & 0.089 & 0.096 & $+$129\% \\
\bottomrule
\end{tabular}
\end{table*}

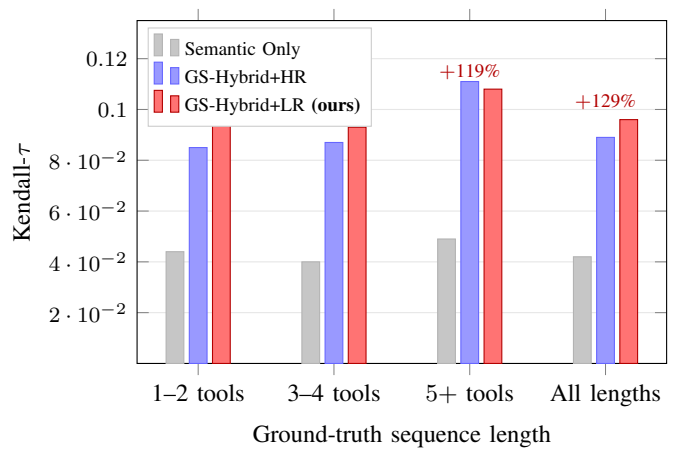
\begin{figure}[!t]
\centering
\resizebox{\columnwidth}{!}{%
\begin{tikzpicture}
\begin{axis}[
  ybar,
  bar width=6.5pt,
  width=8.4cm, height=6.0cm,
  xlabel={Ground-truth sequence length},
  ylabel={Kendall-$\tau$},
  xlabel style={font=\small},
  ylabel style={font=\small},
  tick label style={font=\footnotesize},
  symbolic x coords={1-2 tools, 3-4 tools, 5+ tools, All lengths},
  xtick=data,
  xticklabels={1--2 tools, 3--4 tools, $5{+}$ tools, All lengths},
  xticklabel style={font=\small},
  ymin=0.0, ymax=0.135,
  ytick={0.02,0.04,0.06,0.08,0.10,0.12},
  yticklabel style={font=\footnotesize},
  legend style={font=\scriptsize, at={(0.02,0.98)}, anchor=north west,
    inner sep=2pt, row sep=-1pt, draw=gray!40},
  legend cell align=left,
  ymajorgrids=true,
  grid style={gray!20, line width=0.4pt},
  enlarge x limits=0.15,
]

\addplot[fill=gray!45, draw=gray!60]
  coordinates {
    (1-2 tools,  0.044)
    (3-4 tools,  0.040)
    (5+ tools,   0.049)
    (All lengths,0.042)
  };
\addlegendentry{Semantic Only}

\addplot[fill=blue!40, draw=blue!60]
  coordinates {
    (1-2 tools,  0.085)
    (3-4 tools,  0.087)
    (5+ tools,   0.111)
    (All lengths,0.089)
  };
\addlegendentry{GS-Hybrid+HR}

\addplot[fill=red!55, draw=red!70!black]
  coordinates {
    (1-2 tools,  0.098)
    (3-4 tools,  0.093)
    (5+ tools,   0.108)
    (All lengths,0.096)
  };
\addlegendentry{GS-Hybrid+LR \textbf{(ours)}}

\node[font=\scriptsize, text=red!70!black, anchor=south]
  at (axis cs:1-2 tools,  0.098) {\raisebox{1pt}{$+$123\%}};
\node[font=\scriptsize, text=red!70!black, anchor=south]
  at (axis cs:3-4 tools,  0.093) {\raisebox{1pt}{$+$134\%}};
\node[font=\scriptsize, text=red!70!black, anchor=south]
  at (axis cs:5+ tools,   0.108) {\raisebox{1pt}{$+$119\%}};
\node[font=\scriptsize, text=red!70!black, anchor=south]
  at (axis cs:All lengths,0.096) {\raisebox{1pt}{$+$129\%}};

\end{axis}
\end{tikzpicture}}%
\caption{Kendall-$\tau$ by ground-truth sequence length on ToolBench (9,965
instances). GS-Hybrid+LR (ours) improves over Semantic Only by $+$119--134\%
across \emph{all} length buckets, confirming that SkillGraph dependency priors
are beneficial regardless of workflow length. Percentages above bars indicate
relative improvement of LR over Semantic Only.}
\label{fig:length_bar}
\end{figure}

Table~\ref{tab:length} breaks down Kendall-$\tau$ by sequence length.
LR improves over Semantic Only by $+119\%$ to $+134\%$ across all length buckets---SkillGraph dependency priors help regardless of sequence length.
Even short sequences (1--2 tools) benefit substantially (+123\%): pairwise dependencies exist even in two-tool workflows (e.g., a query tool must precede a format/send tool).
The absolute Kendall-$\tau$ of LR is non-monotone across buckets (0.098 for 1--2 tools, 0.093 for 3--4, 0.108 for 5+), but the relative gain over Semantic Only remains consistent: longer sequences provide more pairwise comparisons, and graph dependency priors are most informative in that regime.
Figure~\ref{fig:length_bar} visualises these results across all length buckets.

\subsection{Pareto Optimality Analysis}

\begin{figure}[!t]
\centering
\begin{tikzpicture}
\begin{axis}[
  xlabel={Set-F1},
  ylabel={Ordered Precision},
  xlabel style={font=\small},
  ylabel style={font=\small},
  tick label style={font=\footnotesize},
  xmin=0.195, xmax=0.288,
  ymin=0.055, ymax=0.172,
  xtick={0.21,0.23,0.25,0.27},
  ytick={0.07,0.09,0.11,0.13,0.15,0.17},
  xticklabel style={font=\footnotesize},
  yticklabel style={font=\footnotesize},
  width=7.8cm, height=6.4cm,
  legend style={font=\footnotesize, at={(0.04,0.96)}, anchor=north west,
    row sep=-1pt, inner sep=3pt},
  legend cell align=left,
  grid=both,
  grid style={gray!20, line width=0.4pt},
  major grid style={gray!35, line width=0.5pt},
]

%% Single-stage baselines (filled circles, blue)
\addplot[only marks, mark=*, mark size=2.8pt,
         draw=blue!70!black, fill=blue!30, opacity=0.9]
  coordinates {
    (0.215, 0.073)   % BM25
    (0.266, 0.123)   % Semantic Only
    (0.240, 0.149)   % Beam Search
    (0.271, 0.130)   % Hybrid Sem-Graph
  };
\addlegendentry{Single-stage}

%% Two-stage, non-learned (filled triangles, orange)
\addplot[only marks, mark=triangle*, mark size=3.2pt,
         draw=orange!80!black, fill=orange!40, opacity=0.9]
  coordinates {
    (0.271, 0.133)   % GS-Hybrid + Sem-Sort
    (0.271, 0.152)   % GS-Hybrid + Hybrid-Rerank
  };
\addlegendentry{Two-stage (non-learned)}

%% Our method (star, red)
\addplot[only marks, mark=star, mark size=5.5pt,
         draw=red!80!black, fill=red!70, opacity=1.0]
  coordinates {
    (0.271, 0.156)   % GS-Hybrid + Learned Reranker (ours)
  };
\addlegendentry{GS-Hybrid+LR \textbf{(ours)}}

%% Point labels
% BM25
\node[font=\scriptsize, text=blue!70!black, anchor=north]
  at (axis cs:0.215,0.073) {BM25};
% Semantic Only
\node[font=\scriptsize, text=blue!70!black, anchor=north east]
  at (axis cs:0.266,0.123) {Sem-Only};
% Beam Search
\node[font=\scriptsize, text=blue!70!black, anchor=south]
  at (axis cs:0.240,0.149) {Beam};
% Hybrid Sem-Graph
\node[font=\scriptsize, text=blue!70!black, anchor=south east]
  at (axis cs:0.271,0.130) {Hybrid};
% Our method
\node[font=\scriptsize, text=red!80!black, anchor=south west]
  at (axis cs:0.271,0.156) {\textbf{Ours}};

%% Pareto-front arrow annotation
\draw[-{Stealth[length=4pt,width=3pt]}, gray!60, dashed]
  (axis cs:0.250,0.105) -- (axis cs:0.268,0.152);
\node[font=\scriptsize, gray!70, align=center]
  at (axis cs:0.243,0.098) {Pareto\\frontier};

\end{axis}
\end{tikzpicture}
\caption{Set-F1 vs.\ Ordered Precision on ToolBench (9,965 instances) for representative methods.
GS-Hybrid+LR (\textcolor{red!80!black}{$\bigstar$}) is the sole
Pareto-optimal point, achieving the highest value on \emph{both} axes
simultaneously.
Single-stage methods (\textcolor{blue!60!black}{$\bullet$}) trade
Set-F1 for Ord.Prec or vice versa.
Two-stage non-learned methods
(\textcolor{orange!80!black}{$\blacktriangle$}) improve ordering without
sacrificing Set-F1, but fall short of the learned reranker.}
\label{fig:pareto}
\end{figure}
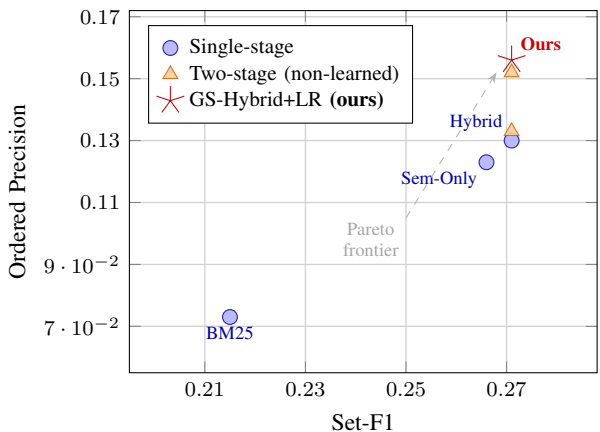

Fig.~\ref{fig:pareto} plots Set-F1 against Ordered Precision for representative methods, revealing the Pareto frontier of the selection-ordering trade-off.
Single-stage methods cluster along a dominated frontier: Beam Search achieves higher Ord.Prec (0.149) than Semantic Only (0.123) but at lower Set-F1 (0.240 vs.\ 0.271), illustrating that joint optimization compromises one objective to improve the other.
Hybrid Sem-Graph improves Set-F1 to 0.271 but its Ord.Prec (0.130) barely exceeds Semantic Only.

Two-stage methods break this trade-off by fixing Set-F1 at 0.271 (the Stage-1 ceiling) while independently maximizing ordering quality.
GS-Hybrid+HR reaches Ord.Prec = 0.152; GS-Hybrid+LR reaches 0.156.
Only GS-Hybrid+LR is Pareto-optimal: no other method achieves equal or better Set-F1 \emph{and} equal or better Ord.Prec simultaneously.
The gap between graph-only Opt-Perm (Ord.Prec = 0.151) and LR (0.156) is small, confirming that learned pairwise scoring extracts signal beyond raw transition likelihood.

\subsection{Error Analysis and First-Tool Accuracy Trade-off}

LR improves global ordering at the cost of First-Tool Accuracy: 1st.Acc drops from 0.262 (Sem-Sort) to 0.234 (LR), a regression of 2.8 percentage points.
To understand when LR helps and hurts, we examine per-sample Ord.Prec changes between LR and Sem-Sort on the full test set (9,965 instances):
515 instances improve ($+$5.2\%), 243 instances degrade ($-$2.4\%), and 9,207 remain unchanged (92.4\%).

Improved cases share a common pattern: they involve pairs of tools with high SkillGraph transition asymmetry (large $|w(t_a, t_b) - w(t_b, t_a)|$), where the semantic scores incorrectly ranked the dependent tool first.
Degraded cases predominantly involve \emph{single-tool} sequences or sequences where all permutations are equally valid (e.g., two independent API calls with no data-flow dependency); in these cases, the pairwise signal is uninformative and LR may arbitrarily reorder.

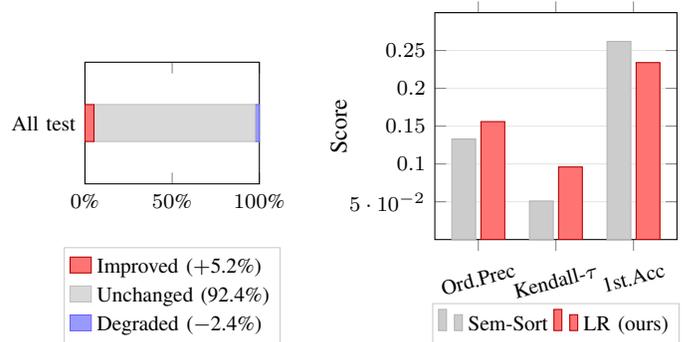
\begin{figure}[!t]
\centering
%% ── Left panel ──────────────────────────────────────────────────────────────
\begin{minipage}[t]{0.44\columnwidth}
\centering
\begin{tikzpicture}
\begin{axis}[
  xbar stacked,
  bar width=14pt,
  width=3.9cm, height=3.2cm,
  xmin=0, xmax=100,
  xtick={0,50,100},
  xticklabel={\pgfmathprintnumber{\tick}\%},
  xticklabel style={font=\footnotesize},
  ytick={0},
  yticklabels={All test},
  yticklabel style={font=\footnotesize},
  ymin=-0.5, ymax=0.5,
  legend style={font=\footnotesize, at={(0.5,-0.52)}, anchor=north,
    legend columns=1, inner sep=2pt, draw=gray!40},
  legend cell align=left,
]
\addplot[fill=red!55, draw=red!70!black]
  coordinates {(5.17, 0)};
\addlegendentry{Improved ($+$5.2\%)}
\addplot[fill=gray!30, draw=gray!50]
  coordinates {(92.43, 0)};
\addlegendentry{Unchanged (92.4\%)}
\addplot[fill=blue!40, draw=blue!60]
  coordinates {(2.44, 0)};
\addlegendentry{Degraded ($-$2.4\%)}
\end{axis}
\end{tikzpicture}
\vspace{2pt}\\
{\footnotesize (a) Per-sample $\Delta$Ord.Prec}
\end{minipage}%
\hfill
%% ── Right panel ─────────────────────────────────────────────────────────────
\begin{minipage}[t]{0.52\columnwidth}
\centering
\begin{tikzpicture}
\begin{axis}[
  ybar,
  bar width=9pt,
  width=4.8cm, height=4.6cm,
  symbolic x coords={OrdPrec, KendallTau, FirstAcc},
  xtick=data,
  xticklabels={Ord.Prec, Kendall-$\tau$, 1st.Acc},
  xticklabel style={font=\footnotesize, rotate=15},
  ymin=0.0, ymax=0.30,
  ytick={0.05,0.10,0.15,0.20,0.25},
  scaled y ticks=false,
  yticklabel style={font=\footnotesize},
  ylabel={Score},
  ylabel style={font=\small, yshift=-6pt},
  legend style={font=\footnotesize, at={(0.5,-0.28)}, anchor=north,
    legend columns=2, inner sep=2pt, draw=gray!40},
  legend cell align=left,
  ymajorgrids=true,
  grid style={gray!20, line width=0.4pt},
  enlarge x limits=0.28,
]
\addplot[fill=gray!40, draw=gray!60]
  coordinates {
    (OrdPrec,    0.133)
    (KendallTau, 0.051)
    (FirstAcc,   0.262)
  };
\addlegendentry{Sem-Sort}
\addplot[fill=red!55, draw=red!70!black]
  coordinates {
    (OrdPrec,    0.156)
    (KendallTau, 0.096)
    (FirstAcc,   0.234)
  };
\addlegendentry{LR (ours)}
\end{axis}
\end{tikzpicture}
\vspace{2pt}\\
{\footnotesize (b) Ordering vs.\ 1st.Acc trade-off}
\end{minipage}
\caption{Error analysis of GS-Hybrid+LR vs.\ GS-Hybrid+Sem-Sort on ToolBench.
(a)~92\% of instances are unaffected; LR improves 515 instances ($+$5.2\%) and
degrades 243 ($-$2.4\%). Improved cases share high SkillGraph transition
asymmetry; degraded cases are predominantly single-tool sequences where
pairwise signal is uninformative.
(b)~LR improves global ordering (Ord.Prec $+17\%$; Kendall-$\tau$ $+88\%$) at
a small cost to first-tool accuracy ($-$2.8\,pp), a known property of pairwise
ranking objectives that optimise aggregate agreement rather than position~1.}
\label{fig:error_analysis}
\end{figure}

The 1st.Acc regression is a known property of global pairwise ranking objectives~\cite{ranknet}: optimizing aggregate pairwise agreement does not explicitly protect the first position.
A practical mitigation---anchoring the highest-semantics-scoring tool to position 1 before applying pairwise reranking---is left as an implementation option for latency-sensitive deployments.
Figure~\ref{fig:error_analysis} summarises both the improvement/degradation breakdown and the global-ordering vs.\ first-position trade-off.

\subsection{Fixed-$K$ Robustness}

\begin{table}[!t]
\caption{Fixed-$K$ Robustness on ToolBench (9,965 test instances).
Semantic~Only uses top-$K$ retrieval; GS-Hybrid+LR uses hybrid
graph-semantic stage~1 truncated to $K$, then the learned reranker.
\textit{Oracle} uses $K=\max(|S^*|,3)$ as the reference.
$\Delta\tau$ = relative improvement in Kendall-$\tau$.}
\label{tab:fixed_k}
\centering
\small
\setlength{\tabcolsep}{4pt}%
\begin{tabular}{llccc}
\toprule
$K$ & \textbf{Method} & \textbf{Set-F1} & \textbf{Kendall-$\tau$} & \textbf{$\Delta\tau$} \\
\midrule
\multirow{2}{*}{3}
 & Semantic Only       & 0.269 & 0.037 & -- \\
 & GS-Hybrid+LR        & \textbf{0.274} & \textbf{0.095} & $+$156\% \\
\midrule
\multirow{2}{*}{5}
 & Semantic Only       & 0.237 & 0.052 & -- \\
 & GS-Hybrid+LR        & \textbf{0.243} & \textbf{0.107} & $+$106\% \\
\midrule
\multirow{2}{*}{8}
 & Semantic Only       & 0.195 & 0.068 & -- \\
 & GS-Hybrid+LR        & \textbf{0.197} & \textbf{0.121} & $+$79\% \\
\midrule
\multirow{2}{*}{Oracle}
 & Semantic Only       & 0.266 & 0.042 & -- \\
 & GS-Hybrid+LR\,(main) & \textbf{0.271} & \textbf{0.096} & $+$129\% \\
\bottomrule
\end{tabular}
\end{table}

The main evaluation uses Oracle-$K$ ($K = \max(|S^*|, 3)$) to isolate selection and ordering quality from sequence-length estimation.
Table~\ref{tab:fixed_k} verifies that the ordering advantage of GS-Hybrid+LR is \emph{not} an artefact of oracle length knowledge: at every fixed $K \in \{3, 5, 8\}$, GS-Hybrid+LR improves Kendall-$\tau$ over Semantic Only by $+$79\% to $+$156\%.
The relative gain is largest at small $K$ (where graph dependencies are most concentrated) and decreases as $K$ grows (more tools dilute the transition signal).
Set-F1 advantage is consistent but modest (+0.5--0.8 pp) across all $K$, confirming that the two-stage decoupling primarily adds ordering value rather than altering set coverage.
These fixed-$K$ results reduce the dependence on oracle length knowledge, but they still do not solve the full deployment problem because $K$ is prescribed globally rather than inferred per query.

Notably, under fixed $K=5$, GS-Hybrid+LR achieves a higher Kendall-$\tau$ (0.107) than the oracle-$K$ setting (0.096), because oracle $K$ includes many single- and two-tool instances (where $\tau$ is trivially low), which suppress the macro-average.
Graph-guided candidate expansion and learned pairwise ordering are thus robust to the choice of $K$---the gains are not an artefact of oracle length knowledge.

% ============================================================
\section{Limitations}
\label{sec:limitations}
% ============================================================

\textbf{Oracle-$K$ evaluation.}
The main results use $K = \max(|S^*|, 3)$, assuming the sequence length is known at test time.
Table~\ref{tab:fixed_k} (Section~\ref{sec:analysis}) demonstrates that the ordering advantage holds across all fixed-$K$ settings ($K \in \{3,5,8\}$), with Kendall-$\tau$ improvements of $+$79--156\%.
This reduces the concern that our gains are tied to one specific oracle length, but it does not eliminate the deployment gap: predicting $K$ automatically from the query remains unsolved in this work and may interact with both selection and ordering quality.

\textbf{API-Bank sample size.}
The API-Bank Level-3 benchmark contains only 50 instances, with mean sequence length 2.28 and maximum length 3, limiting both statistical power and the diversity of workflow structures probed by the validation.
The Kendall-$\tau$ improvement (+1.046 absolute over the semantic-only baseline) is large, but these results are best treated as a low-resource validation until confirmed on larger, longer structured-workflow benchmarks.

\textbf{Offline evaluation.}
Our experiments measure recommendation quality against historical trajectories.
Real agent performance depends on downstream execution success, which may diverge from trajectory-matching metrics.

\textbf{First-Tool Accuracy regression.}
The Learned Reranker improves global ordering at the cost of first-tool accuracy (0.234 vs. 0.262 for Sem-Sort).
Applications requiring a precise first tool may prefer a hybrid that preserves first-position scores.

\textbf{Cold-start tools.}
SkillGraph edges are derived from co-occurrence in training trajectories.
In the processed ToolBench test split, 110 of 1,317 unique test-side tools are absent from the training graph (91.6\% coverage), and even covered tools may have very sparse transition counts.
For these cold-start or rare tools, the method must fall back more heavily to semantic scoring, so the current graph prior is strongest for tools with at least moderate historical support.

% ============================================================
\section{Conclusion}
\label{sec:conclusion}
% ============================================================

We presented SkillGraph, a directed weighted graph of tool execution-transition regularities mined from large-scale LLM agent trajectories, and a two-stage decoupled framework for tool sequence recommendation.
Our central finding is the \emph{selection-ordering signal gap}: semantic similarity is an adequate proxy for tool selection but systematically harmful for ordering, producing negative Kendall-$\tau$ in structured workflow domains where inter-tool dependencies govern execution order.
SkillGraph makes these precedence cues explicit as a graph foundation prior---constructed once from trajectory data and reused across queries---providing the ordering signal that embeddings cannot encode.

The two-stage framework acts on this directly: hybrid graph-semantic retrieval for selection, and a learned pairwise reranker with SkillGraph features for ordering.
The result is Pareto-optimal performance on ToolBench (9,965 instances, $\sim$16,000 tools) and a dramatic low-resource improvement on API-Bank (Kendall-$\tau$: $-0.433 \to +0.613$), while remaining computationally lightweight---outperforming LLaMA-3.1-8B rerankers with a 3-layer MLP trained in under five minutes.

For structured, dependency-driven planning tasks, purpose-built graph priors from empirical data outperform large parametric language models in our setting.
Tool execution dependencies are not captured by semantic representations and are not reliably inferred by LLMs; the best proxy comes from successful executions.
Trajectory-mined dependency graphs are a practical foundation for LLM agent planning at scale---our cross-dataset results provide early evidence of this.

% ============================================================
% REFERENCES
% ============================================================
\bibliographystyle{IEEEtran}
\bibliography{IEEEabrv,main}

\end{document}